\PassOptionsToPackage{table,xcdraw}{xcolor}
\documentclass{article}

\usepackage{microtype}
\usepackage{graphicx}
\usepackage{booktabs} % for professional tables

\usepackage{subcaption}
\usepackage{enumitem} % Add this in your preamble

\usepackage{hyperref}

\usepackage[accepted]{icml2025}

\usepackage{amsmath}
\usepackage{amssymb}
\usepackage{mathtools}
\usepackage{amsthm}

\usepackage{capt-of,etoolbox}
\usepackage{multirow}
\usepackage{fontawesome5}
\usepackage[normalem]{ulem}
\useunder{\uline}{\ul}{}
\usepackage{amsmath} % for using \text command

\usepackage{tikz} % for drawing the circle
\usepackage{bbm}

\usepackage[font=small,skip=4pt]{caption} % or tweak skip as needed
\usepackage[capitalize,noabbrev]{cleveref}

\theoremstyle{plain}

\theoremstyle{definition}

\theoremstyle{remark}

\usepackage[textsize=tiny]{todonotes}

\def\ours{EgoPrivacy}

\usepackage{pifont}% http://ctan.org/pkg/pifont

\newcommand{\cmark}{\color[HTML]{009B55} \ding{51}}%
\newcommand{\xmark}{\color{red} \ding{55}}%

\title{\ours{}: What Your First-Person Camera Says About You?}
\icmltitlerunning{\ours{}: What Your First-Person Camera Says About You?}

\begin{document}

\twocolumn[{
\icmltitle{\ours{}: What Your First-Person Camera Says About You?}

% It is OKAY to include author information, even for blind
% submissions: the style file will automatically remove it for you
% unless you've provided the [accepted] option to the icml2025
% package.

% List of affiliations: The first argument should be a (short)
% identifier you will use later to specify author affiliations
% Academic affiliations should list Department, University, City, Region, Country
% Industry affiliations should list Company, City, Region, Country

% You can specify symbols, otherwise they are numbered in order.
% Ideally, you should not use this facility. Affiliations will be numbered
% in order of appearance and this is the preferred way.
\icmlsetsymbol{equal}{*}

\begin{icmlauthorlist}
\icmlauthor{Yijiang Li}{ucsd}
\icmlauthor{Genpei Zhang}{uestc}
\icmlauthor{Jiacheng Cheng}{ucsd}
\icmlauthor{Yi Li}{qti}
\icmlauthor{Xiaojun Shan}{ucsd}
\icmlauthor{Dashan Gao}{qti}
\icmlauthor{Jiancheng Lyu}{qti}
%\icmlauthor{}{sch}
\icmlauthor{Yuan Li}{qti}
\icmlauthor{Ning Bi}{qti}
\icmlauthor{Nuno Vasconcelos}{ucsd}
%\icmlauthor{}{sch}
%\icmlauthor{}{sch}
\end{icmlauthorlist}

\icmlaffiliation{ucsd}{University of California, San Diego}
\icmlaffiliation{qti}{Qualcomm AI Research, an initiative of Qualcomm Technologies, Inc}
% \icmlaffiliation{qti}{Qualcomm AI Research}
\icmlaffiliation{uestc}{ University of Electronic Science and Technology of China}

\icmlcorrespondingauthor{Jiacheng Cheng}{jiacheng.cheng96@gmail.com}
% \icmlcorrespondingauthor{Firstname2 Lastname2}{first2.last2@www.uk}

% You may provide any keywords that you
% find helpful for describing your paper; these are used to populate
% the "keywords" metadata in the PDF but will not be shown in the document
\icmlkeywords{egocentric vision, privacy, egocentric video, privacy attack, benchmark}
% \begin{center}
%     \centering
%     \includegraphics[width=0.8\linewidth]{assets/teaser_v1.pdf}
%     \\
%     \captionof{figure}{\small \textbf{Overview of the proposed \ours{} benchmark.} {What can you tell about the camera wearer from egocentric videos alone?}
%     It may come as a surprise that a fair amount of information about the user, such as demographics, identity, time and location of recording, can be inferred from their first-person view footages, despite not revealing their faces or full body.}%
%     \label{fig:teaser}
%     % \vspace{1em}
% \end{center}

\vskip 0.3in
}]

% this must go after the closing bracket ] following \twocolumn[ ...

% This command actually creates the footnote in the first column
% listing the affiliations and the copyright notice.
% The command takes one argument, which is text to display at the start of the footnote.
% The \icmlEqualContribution command is standard text for equal contribution.
% Remove it (just {}) if you do not need this facility.

%\printAffiliationsAndNotice{}  % leave blank if no need to mention equal contribution
% \printAffiliationsAndNotice{\icmlEqualContribution} % otherwise use the standard text.
\printAffiliationsAndNotice{} % otherwise use the standard text.

% \begin{abstract}
% This document provides a basic paper template and submission guidelines.
% Abstracts must be a single paragraph, ideally between 4--6 sentences long.
% Gross violations will trigger corrections at the camera-ready phase.
% \end{abstract}

\begin{abstract}
While the rapid proliferation of wearable cameras has raised significant concerns about egocentric video privacy, prior work has largely overlooked the unique privacy threats posed to the camera wearer. This work investigates the core question: \emph{How much privacy information about the camera wearer can be inferred from their first-person view videos?} We introduce EgoPrivacy, the first large-scale benchmark for comprehensive evaluation of privacy risks in egocentric vision. EgoPrivacy covers three types of privacy (demographic, individual, and situational) defining seven tasks that aim to recover private information ranging from fine-grained (e.g., wearer's identity) to coarse-grained (e.g., age group). To further emphasize the privacy threats inherent to egocentric vision, we propose \emph{Retrieval-Augmented Attack}, a novel attack strategy that leverages ego-to-exo retrieval from an external pool of exocentric videos to boost the effectiveness of demographic privacy attacks. An extensive comparison of the different attacks possible under all threat models is presented, showing that private information of the wearer is highly susceptible to leakage. For instance, our findings indicate that foundation models can effectively compromise wearer privacy even in zero-shot settings by recovering attributes such as identity, scene, gender, and race with 70–80\% accuracy. Our code and data are available at \url{https://github.com/williamium3000/ego-privacy}.

\end{abstract}    
\begin{figure*}
    \centering
    \includegraphics[width=0.8\linewidth]{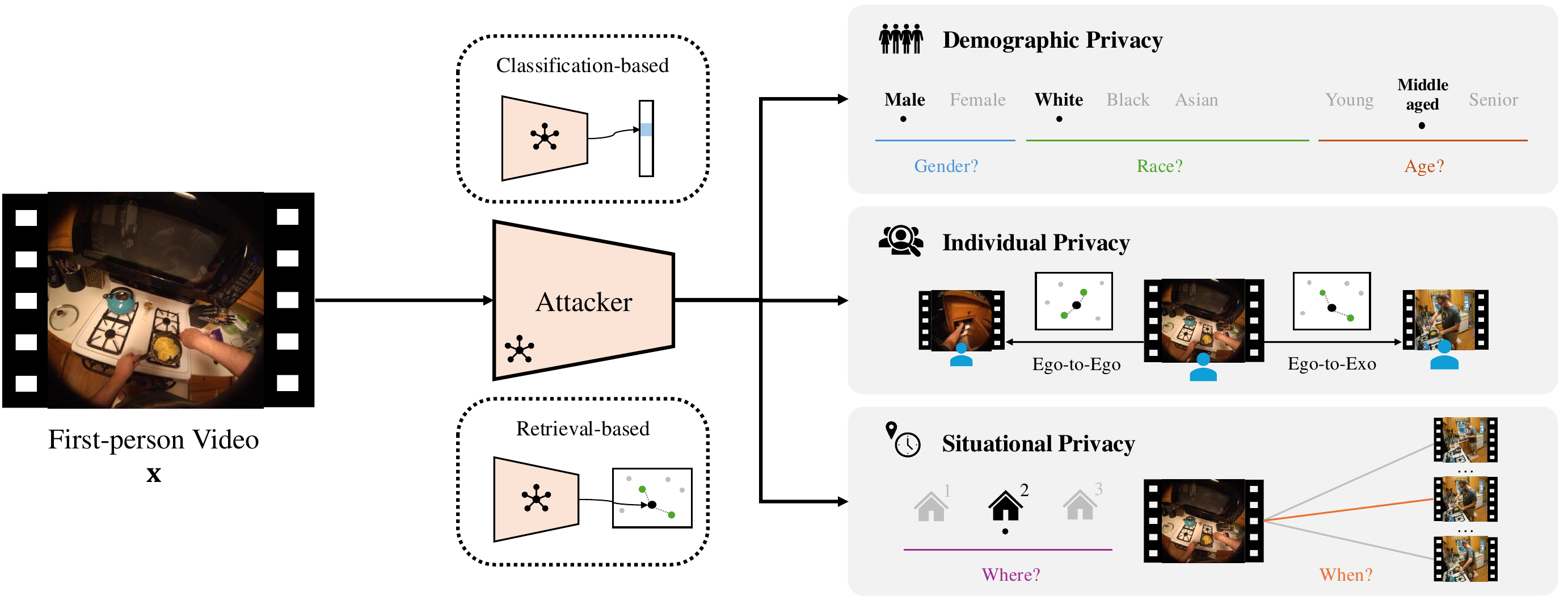}
    \\
        \vspace{-0.1in}
    \captionof{figure}{\small \textbf{Overview of the proposed \ours{} benchmark.} {What can you tell about the camera wearer from egocentric videos alone?}
    It may come as a surprise that a fair amount of information about the user, such as demographics, identity, time and location of recording, can be inferred from their first-person view footages, despite not revealing their faces or full body.}%
    \label{fig:teaser}
    % \vspace{1em}
    % \vspace{-0.1in}
\end{figure*}

\section{Introduction}
\label{sec:intro}
% Wearable cameras and egocentric (first-person view)  videos offer many important advantages for applications like augmented reality and assistive technology for the visually impaired. As advancements in hardware continue to enhance affordability, miniaturization, durability, and energy efficiency, the adoption of these technologies is becoming increasingly widespread~\cite{betancourt2015evolution,plizzari2024outlook}. Computer vision promises to magnify their utility, enabling applications such as activity recognition~\cite{nguyen2016recognition}, human behavior analysis~\cite{cazzato2020look}, or life logging~\cite{bolanos2016toward,del2016summarization}. Unsurprisingly, large datasets of egocentric video are starting to appear~\cite{sigurdsson2018actor,Grauman_2022_CVPR,grauman2024ego} and research in the topic is increasing. The resulting new applications will increase the size of the egocentric video market, encouraging further progress in the development of cheaper and easier to use cameras, in turn enabling more sophisticated computer vision applications and connections to AI systems that analyze video for all sorts of reasons. While beneficial for many applications, this positive feedback loop raises unprecedented concerns about privacy~\cite{hoyle2014privacy,hoyle2015sensitive}. 

The growing adoption of wearable cameras and egocentric (first-person view) videos, driven by advances in hardware and computer vision ~\cite{betancourt2015evolution,plizzari2024outlook, sigurdsson2018actor,Grauman_2022_CVPR,grauman2024ego}, enables innovative applications like activity recognition~\cite{nguyen2016recognition}, human behavior analysis~\cite{cazzato2020look}, or life logging~\cite{bolanos2016toward,del2016summarization}. However, it also raises significant privacy concerns~\cite{hoyle2014privacy,hoyle2015sensitive}. An already popular concern is the privacy of people \emph{captured by} egocentric cameras~\cite{farringdon2000visual,krishna2005wearable,mandal2015wearable,chakraborty2016person,templeman2014placeavoider,korayem2016enhancing,dimiccoli2018mitigating,hasan2017cartooning,fergnani2016body}. This concern, however, is not specific to egocentric video. Third-person cameras are already common in public environments, e.g. surveillance networks, and many private environments, e.g. TV sets with user facing cameras, motivating  a line of research on privacy preserving cameras~\cite{hinojosa2021learning,hinojosa2022privhar,cheng2024learning,khan2024opencam} and post-hoc privacy techniques, e.g. methods to delete or obfuscate faces in images~\cite{criminisi2003object,criminisi2004region,bitouk2008face,ren2018learning}. 
% This has motivated a literature on privacy preserving cameras~\cite{hinojosa2021learning,hinojosa2022privhar,cheng2024learning}, which collect images that support computer vision tasks but are not understandable enough to allow recognition of people or objects. There is also a literature on post-hoc privacy techniques, e.g. methods to delete or obfuscate faces in images~\cite{criminisi2003object,criminisi2004region,bitouk2008face,ren2018learning}. 
While sharing all these issues, egocentric video introduces a new set of privacy concerns of its own, namely the privacy implications for the camera \emph{wearers}, which have been much less studied~\cite{hoshen2016egocentric,thapar2020sharing,thapar2020recognizing,tsutsui2021whose}.

{\it Wearer-centric\/} privacy is particularly concerning because egocentric videos are highly personal, captured continuously to document the day-to-day experience and surroundings of the camera wearer, and to keep track of their activities~\cite{plizzari2024outlook}. The availability of this information will create pressures for its sharing, e.g. free video storage in exchange for video mining access, analysis by third parties, e.g. insurance companies collecting health information, and cross-referencing of egovideo with publicly available third-person video of the wearer, e.g. on social media platforms. All privacy problems currently posed by location-tracking apps will be magnified by the ability to know not only where people are but also {\it what they are doing\/}~\cite{hoyle2014privacy, price2017logging, 8953346}. All of this can lurk under a false sense of privacy,  due to the fact that the camera is not framing its user. Given the limited attention to the problem, it is currently not even well understood {\it how much\/} of a privacy problem egocentric video poses to camera wearers. Questions such as what type of private information and how much of it can be recovered remain largely unanswered. 

%However, in the absence of high-quality data and well-defined metrics, there has been limited progress toward systematically understanding and measuring these privacy risks, let alone how to mitigate them.

This work is a first attempt to define the range of \emph{wearer-centric} privacy problems arising from egocentric recordings. In essence, we ask: \emph{What can be told about the camera wearer by watching egocentric videos?} \cref{fig:teaser} illustrates a variety of personal information that can be inferred from the video: hand appearance and pose can give away the gender, race and age of the wearer; egocentric videos can be matched to exocentric views of the wearer  to fully reveal identity or activities; background settings and objects can give away location and activity; video clips can be matched to reason about location and time, and so forth. We group these privacy issues into three broad categories: \emph{demographic} privacy for recognizing demographic groups of the wearer, \emph{individual} privacy for uniquely identifying the wearer, and \emph{situational} privacy for recognizing when and where the recording took place. 

To comprehensively study the problem of egovideo privacy, we propose a novel large-scale benchmark, \textbf{\ours{}}, annotated to allow the quantification of privacy risks under each of these categories. \ours{} covers seven tasks representative of the three privacy categories, each formulated as either a problem of video classification or retrieval. We then propose a set of threat models with increasing levels of access to wearer data and perform an extensive evaluation of their ability to recover private information, using various types of foundation models.

Extensive experiments reveal significant privacy challenges, as {\it all\/} threat models are able to extract surprisingly high amounts of private information. For example, zero-shot foundation models are shown to have a remarkable ability to compromise demographic privacy. This implies that even an adversary with no additional data or information about the wearer, can simply use open source models to recover attributes like race and gender. Fine-tuning these models on annotated exocentric or egocentric datasets extends this ability to recover attributes like wearer identity or scene location.

The gap between privacy attacks on egocentric and exocentric video largely owes to a key advantage of egocentric footage: it naturally hides the wearer’s face and most parts of the body which can easily give away the privacy information of a subject. However, in practice, as almost everyone is increasingly exposed to all kinds of cameras in public, it is entirely possible that the camera wearer of an exocentric video will also be filmed in exocentric videos by a third part (e.g.\ suveilance systems, vloggers) simultaneously.  If an adversary could get access to a repository of third-person view videos and successfully recover those third-person view corresponding to the ego video query, the risk of privacy leakage in egocentric vision will be elevated another level. Motivated by this, we introduce the novel \emph{Retrieval-Augmented Attack} (RAA): With access to a repository of third-person videos that may feature the target user, an attacker first conducts ego-to-exo retrieval, then launches the privacy attack from the exocentric perspective. Experiments show that merging cues from the egocentric stream with the retrieved exocentric clip markedly raises the success rate of demographic-privacy attacks.

The gap between privacy attacks on egocentric and exocentric video can be attributed to a key advantage of egocentric footage: it naturally obscures the wearer’s face and much of their body, that typically reveal private information. However, in practice, individuals are increasingly exposed to various public-facing cameras, making it highly plausible that the wearer of an egocentric camera is simultaneously captured in third-person view footages, e.g.\ by surveillance systems or bystanders recording with personal devices. This scenario is far from hypothetical. For instance, consider a case where someone uploads a series of egocentric videos to social media. An attacker could potentially obtain the poster’s IP address and retrieve surveillance footage from nearby locations. Motivated by this, we propose a novel Retrieval-Augmented Attack (RAA): the adversary first performs ego-to-exo retrieval to identify third-person clips containing the target, then launches a privacy attack from the exocentric perspective.  Our experiments demonstrate that incorporating cues from retrieved third-person views into the analysis of egocentric footage significantly improves the effectiveness of demographic privacy attacks.

Overall, this paper makes four key contributions. First, we develop the first comprehensive large-scale benchmark for studying privacy in egocentric videos, which covers risks at the demographic, individual, and situational levels. Second, we formulate various threat models based on attacks with varying levels of access to video of the wearer and  instantiate concrete attacker models for each of them. Third, we present an empirical analysis of the success of these attacks, revealing that even the use of zero-shot foundation models can suffice to expose significant amounts of private information. Last but not least, we further derive a novel privacy attack by ego-to-exo retrieval augmentation and demonstrate its effectiveness at exposing demographic attributes. We hope that our work can lay the foundation for future investigations into both offensive and defensive strategies concerning egocentric privacy.

\section{Related Works}
\label{sec:related-work}

\paragraph{Visual Privacy Benchmarks.} Large-scale public benchmarks are indispensable for successful computer vision research. Multiple benchmarks with privacy annotations (e.g.\ PIPA~\cite{zhang2015beyond}, VISPR~\cite{orekondy17iccv}, VizWiz-Priv~\cite{gurari2019vizwiz}) have been established, but their source data are mostly social media images (e.g.\ Twitter), not egocentric. Some egocentric video datasets with wearer identity annotations (e.g.\ FPSI~\cite{fathi2012social}, EVPR~\cite{hoshen2016egocentric}, IITMD~\cite{thapar2020sharing}) can be employed for wearer identification evaluation, but their potential is limited by the insufficient participants and scene diversity.

\paragraph{Privacy Preservation in Egocentric Vision.} 
% Researchers have made attempts to attenuate the privacy risk for egocentric videos from different perspectives. 
A straightforward solution is to disable the camera when sensitive information are detected~\cite{templeman2014placeavoider,korayem2016enhancing}. Beyond this, a line of work proposes to redact sensitive information in an egocentric video using processing techniques such as image degradation~\cite{dimiccoli2018mitigating}, object replacement~\cite{hasan2017cartooning}, and anonymization transformation~\cite{thapar2021anonymizing}. Another line of work investigates how to perform utility tasks with privacy-preserving representation of the egocentric videos/images (e.g.\ extremely downsampled video~\cite{ryoo2017privacy},  text description~\cite{qiu2023egocentric}) instead of the raw RGB data. Despite abundant research, they primarily focus on third-person subjects appearing in egocentric videos. Our work distinguishes itself from them by taking a new perspective, i.e.\ privacy concerns around the camera wearer.

\paragraph{Egocentric Person Identification.}
Person identification has been well-studied in third-person video settings but remains less explored in egocentric scenarios, where the subject can be either individuals in the camera’s field of view or the camera wearer. For the former, the identification usually relies patterns of the face~\cite{farringdon2000visual,krishna2005wearable,mandal2015wearable,chakraborty2016person} or body part~\cite{fergnani2016body}. The identification of the wearer typically depends on head motion signature~\cite{hoshen2016egocentric,thapar2020sharing}, hand gesture~\cite{thapar2020recognizing,tsutsui2021whose}, and photographer style~\cite{thomas2016seeing}.  Some cross-view wearer identification approaches are proposed with additional third-person view~\cite{yonetani2015ego,poleg2015head,zhao2024fusing} or top-view videos~\cite{ardeshir2018integrating,ardeshir2018egocentric}  as auxiliary data.

\paragraph{Relationship Between Egocentric and Exocentric Videos.} The relationship between egocentric and exocentric videos has been investigated in applications such as knowledge transfer~\cite{li2021ego}, cross-view  generation/translation~\cite{liu2020exocentric,liu2021cross,luo2024intention,luo2024put} and retrieval~\cite{elfeki2018third,yu2020first,xu2024retrieval}. The application of cross-view retrieval to the wearer privacy attack has yet to be thoroughly investigated. 

\section{Benchmarking Privacy in First-Person View}
\label{sec:bench}
Most privacy-preserving vision addresses \emph{third-person} video, equating privacy to (in)ability to recognize faces or other features that reveal personal information, like addresses or phone numbers. While this is concerning for egocentric videos, it fails to capture the full range of privacy risks posed by the latter, which can also expose information about the camera wearer's identity, demographics, and surroundings. To address this problem, we propose \textbf{\ours{}}, a multidimensional privacy benchmark for egocentric vision.

\begin{table*}[]
    \scriptsize
    \centering
    \begin{tabular}{cccccccc}
    \toprule
        Benchmark & Modality & \#Subjects  & \#Scenes & Identity &  Demographics & OOD Data\\
        \midrule
        FPSI~\cite{fathi2012social} & Ego  & 6 & \xmark & \cmark & \xmark & \xmark\\
        EVPR~\cite{hoshen2016egocentric} & Ego & 32 & \xmark  & \cmark & \xmark & \xmark\\
        IITMD-WFP~\cite{thapar2020sharing} & Ego & 31 & \xmark & \cmark & \xmark & \xmark\\
        IITMD-WTP~\cite{thapar2020sharing} & Ego+Exo & 12 &  \xmark & \cmark & \xmark & \xmark\\
        \midrule
        \ours{} (Ours) & Ego+Exo & 819 & 131 & \cmark & \cmark & \cmark\\
        \bottomrule
    \end{tabular}
    \captionof{table}{Comparison of existing egocentric privacy benchmarks.}
    \label{tab:my_label}
\end{table*}

\subsection{Privacy Definition}
\label{privacy_definition}
We consider three types of privacy information and their potential of leakage in egocentric videos.
\paragraph{Demographic privacy.}
These attacks aim to recover demographic groups to which the camera wearer belongs. We consider three such groups: gender, race, and age. While not fully identifying a person, these attributes can be leveraged to build user profiles for unwanted solicitation, e.g. targeted advertising, or discriminatory practices, e.g.\ misuse of race or gender information within health applications~\cite{10.1145/2702123.2702183, price2017logging}. 
Since they are categorical variables, we formulate demographic attacks as \emph{classification} problems, where a predictor $f(\cdot)$ aims to infer a demographic attribute  $a$ (e.g.\ \emph{gender}, \emph{race}, and \emph{age})  of the camera wearer from egocentric video $\mathbf{x}$. This is illustrated in Figure~\ref{fig:teaser}. Privacy risk is measured by the  demographic attribute classification accuracy
\begin{equation}
    \textrm{Acc}(\mathcal{D}; f) = \frac{1}{\lvert\mathcal{D}\rvert} \sum_{(\mathbf{x}, a) \in \mathcal{D}} \mathbbm{1}[f(\mathbf{x}) = a],
    \label{eqn:classify}
\end{equation}
where $\mathbbm{1}[{\cdot}]$ is the indicator function. Higher $\textrm{Acc}(\mathcal{D}; f)$ indicates that dataset $\mathcal{D}$ is more vulnerable to privacy attacks.

\paragraph{Individual Privacy.}
These attacks directly aim to recover the camera wearer \emph{identity} $I$. As shown in Figure~\ref{fig:teaser}, this is formulated as a {\it retrieval problem\/}. A latent embedding is first learned, and a retrieval operation is performed to identify the nearest neighbors of the query $\mathbf{x}$. \ours{} considers both the settings where the retrieved video is ego or exocentric. 
%If effective, the availability of identity information in a single egocentric sample (e.g.\ identifying metadata or even an inadvertent face reflection in a mirror) compromise the privacy of all other videos of the wearer.
Privacy risk is measured by the \emph{hit rate} at $k$ (HR$@k$) for retrieval of videos from the wearer of query $\mathbf{x}$
\begin{equation}
    \textrm{HR}@k(\mathcal{D}; g) = \frac{1}{\lvert\mathcal{D}\rvert} \sum_{(\mathbf{x}, I) \in \mathcal{D}} \mathbbm{1}[g^k(\mathbf{x}) \cap \mathcal{T}_I, \neq \emptyset]
    \label{eqn:retrieval}
\end{equation}
where $g$ is the retrieval operator, $g^k(\mathbf{x})$ the top-$k$ retrieved videos and $\mathcal{T}_I$ the set of {\it videos of identity\/} $I$ (the wearer) in dataset $\mathcal{D}$. Depending on the composition of the retrieval set $\mathcal{D}$, we further categorize the Individual Privacy into two tasks. If the retrieved videos are egocentric, the problem is formulated as ego-to-ego retrieval, where both the query $g^k(\mathbf{x})$ and the retrieval set $\mathcal{D}$ consist solely of egocentric videos. Conversely, if the retrieved videos are exocentric, the task becomes ego-to-exo retrieval, where given an egocentric query $g^k(\mathbf{x})$, the goal is to retrieve the exocentric videos from $\mathcal{D}$ with the same identity.

% \subsection{Situational privacy}
\paragraph{Situational privacy.}
Centering on situational awareness, these attacks aim to determine \emph{where} or \emph{when} an egocentric video clip was recorded. We consider two tasks: \emph{scene} and \emph{moment retrieval}. \emph{Scene retrieval} is motivated by the fact that because egocentric videos depict scenes similarly to exocentric videos, they have a similar risk of exposing private scene information~\cite{chen2024scene}. \emph{scene retrieval} seeks to identify the location where the egocentric video was captured. Conversely, \emph{moment retrieval} considers both, location (\emph{where}) and the timing (\emph{when}) of the footage, striving to pinpoint a precise moment in a corresponding exocentric clip, e.g. a clip captured by a different camera~\cite{liu2024towards,luo2024zero}. As illustrated in Figure~\ref{fig:teaser}, both types of privacy are formulated as {\it retrieval problems\/} and evaluated with \eqref{eqn:retrieval}. \emph{Scene retrieval} replaces $\mathcal{T}_I$ with $\mathcal{T}_S$, the set of video clips from $\cal D$ that are recorded in the scene of the query. For \emph{moment retrieval}, $\mathcal{T}_I$ is replaced by $\mathcal{T}$, the set of exocentric video clips from $\cal D$ that are synchronized with the query video, e.g. footage from different third-person camera perspectives.

\subsection{Benchmark Design}
\label{sec:benchmark}
We provide a brief description of the \ours{} benchmark here, further details on the datasets and annotation process can be found in \cref{app:dataset}.
\ours{} is a benchmark of synchronized ego-exo video, built upon Ego-Exo4D~\cite{grauman2024ego} and Charades-Ego~\cite{sigurdsson2018actor}\footnote{All datasets used in the paper were solely downloaded and evaluated by UC San Diego.}. It includes high-quality annotations for the three privacy categories discussed above: demographic labels (gender, age, and race) for each participant, as well as scene and identity annotations for each egocentric video clip. 
\ours{} is composed of 5,625 video clips from Ego-Exo4D, captured by 839 diverse participants across 131 distinct scenes, and 4,000 clips of daily indoor activities from Charades-Ego, recorded by 112 participants in their homes.

All Ego-Exo4D and Charades-Ego clips include time-synchronized egocentric and exocentric videos along with identity annotations for each clip. However, demographic annotations are sparse since they are self-reported by camera wearers, and many were not collected. We leveraged the availability of exocentric videos to manually annotate the demographics of all participants. Camera wearer race, gender, and age labels were collected for all clips using Amazon Mechanical Turk. The label sets of the privacy classification problems were defined to reflect the make-up of the dataset. Gender classes are $\{\textit{Female, Male}\}$\footnote{We note that these are perceived gender classes by the annotators}, Race's are  $\{\textit{Asian, Black, White}\}$\footnote{Other racial categories were omitted due to the low representation in the dataset.}, Age's are $ \{\textit{Young, Middle-aged, Senior}\}$. For individual and situational privacy, we utilize the provided identity and scene annotations from the datasets. For moment retrieval, the location and timing labels are approximated based on clip footage, where each clip is treated as a distinct space-time instance.

The combination of videos from Ego-Exo and Charades-Ego facilitates the formulation of in-distribution (ID) and out-of-distribution (OOD) problem evaluations. Following the train/test split proposed in \cite{grauman2024ego}, we split the Ego-Exo4D videos into a training set ${\cal D}_{train}$, that can be used for model finetuning, and a test set ${\cal D}_{test}$  for ID evaluation. Charades-Ego is then solely used as a test set for OOD evaluation.

Table~\ref{tab:my_label} compares \ours{} with previous egocentric privacy benchmarks~\cite{fathi2012social, hoshen2016egocentric, thapar2020sharing}, which are significantly smaller, focus solely on identity privacy, lack scene and demographic annotations, do not support OOD testing, and primarily consist of egocentric video data.

\section{Egocentric Privacy Attack}
In this section, we will propose our privacy attack to investigate the privacy concern of camera wearer in first-person views. We start by defining a set of threat models in Section \ref{threat_model} and then propose the attacker models in \ref{attack_model}.

\subsection{Attack Capability}
\label{threat_model}
We consider an adversary with the goal of obtaining one of the 7 types of privacy information of the camera wearer from an egocentric query video $\mathbf{x}$. We delineate a spectrum of capabilities ranging from minimal to extensive. 

\noindent{\bf Capability \tikz[baseline=(char.base)]{\node[shape=circle,draw,inner sep=0.5pt] (char) {\textbf{1}}} ({\it zero-shot}):} The adversary has no access to training data. This is the simplest class of attack, implementable by anyone with access to a foundation model.

\noindent{\bf Capability \tikz[baseline=(char.base)]{\node[shape=circle,draw,inner sep=0.5pt] (char) {\textbf{2}}} ({\it fine-tuned}):} The adversary has access to a \emph{labeled training dataset} ${\cal D}_{\text{train}}$ to fine-tune the model for attack purposes. ${\cal D}_{\text{train}}$ can include either egocentric videos, if ${\cal D}_{\text{test}}$ is egocentric, exocentric videos, if ${\cal D}_{\text{test}}$ is exocentric, or both in the case of moment and ego-to-exo identity retrieval.

\noindent{\bf Capability \tikz[baseline=(char.base)]{
            \node[shape=circle,draw,inner sep=0.5pt] (char) {\textbf{3}};} ({\it retrieval-augmented}):} The adversary has access to an identity labeled ego-exo paired training set (for ego-to-exo identity retriever) and an external pool of unlabeled exocentric videos ${\cal D}_{\text{retr}}$, which potentially includes the \emph{identity} of the target egocentric query video $\mathbf{x}$. 

\noindent{\bf Capability \tikz[baseline=(char.base)]{
            \node[shape=circle,draw,inner sep=0.5pt] (char) {\textbf{4}};}({\it identity-level attack}):}
In addition to the capabilities above, the adversary further ascertains whether two egocentric videos share the \emph{same identity}, without necessarily identifying the individuals depicted.

We justify \textbf{Capability \tikz[baseline=(char.base)]{
            \node[shape=circle,draw,inner sep=0.5pt] (char) {\textbf{3}};}} and \textbf{Capability \tikz[baseline=(char.base)]{
            \node[shape=circle,draw,inner sep=0.5pt] (char) {\textbf{4}};}} in \cref{justification}, by outlining realistic threat scenarios in which they arise.
% We additionally discuss and analyze a minor \textbf{Capability \tikz[baseline=(char.base)]{
%             \node[shape=circle,draw,inner sep=0.5pt] (char) {\textbf{4}};}} in \cref{ability4} and justify all of the above capabilities in \cref{justification}, by outlining realistic threat scenarios in which they arise.

% Depending on the private information sought, the attack is formulated as a problem of 1) video \emph{classification} into certain demographic or scene attribute classes, or 2) of \emph{retrieval} of other egocentric or exocentric videos associated with the same user.  While the adversary has no direct access to exocentric videos paired with $\mathbf{x}$, it can utilize an external pool of videos to perform ego-to-exo retrieval and potentially identify the user in the third-person view. Privacy is measured by evaluating classification or retrieval accuracy over a dataset $\cal D$ representative of a target application.
\subsection{Implementation}
\label{attack_model}
In this section, we discuss the implementation of the threat models with different capabilities for each of the three privacy categories.

\noindent\textbf{Demographic Privacy}
% Demographic privacy involves three categories of private information regarding the camera wearer, i.e. gender, race and age. As discussed in Section \ref{privacy_definition}, demographic privacy is modeled as a classification problem where the attacker model is a classifier $f(\cdot)$ with the goal of classifying the correct category of the demographic attribute.
 is modeled as a classification problem, as discussed in Section \ref{privacy_definition}. Here, the classifier $f(\cdot)$ is implemented with a multi-modal foundation model. Capability {\bf \tikz[baseline=(char.base)]{\node[shape=circle,draw,inner sep=0.5pt] (char) {\textbf{1}}}}: $f(\cdot)$ is applied to ${\cal D}_{test}$ in a zero-shot manner.  Capability {\bf \tikz[baseline=(char.base)]{\node[shape=circle,draw,inner sep=0.5pt] (char) {\textbf{2}}}}:$f(\cdot)$ is finetuned on ${\cal D}_{\text{train}}$ and tested on ${\cal D}_{test}$. We consider the {\it in-distribution} (ID), i.e.\  both ${\cal D}_{\text{train}}$  and ${\cal D}_{\text{test}}$ are from Ego-Exo4D and 
the {\it out-of-distribution} (OOD) where ${\cal D}_{train}$ are from Ego-Exo4D and ${\cal D}_{\text{test}}$ from Charades-Ego. For  the combination of capability {\bf \tikz[baseline=(char.base)]{\node[shape=circle,draw,inner sep=0.5pt] (char) {\textbf{1}}}} /  {\bf \tikz[baseline=(char.base)]{\node[shape=circle,draw,inner sep=0.5pt] (char) {\textbf{2}}}} and the additional {\bf \tikz[baseline=(char.base)]{\node[shape=circle,draw,inner sep=0.5pt] (char) {\textbf{3}}}}, both query $\mathbf{x}$ and retrieval dataset ${\cal D}_{\text{retr}}$ are fed to the identity retriever to obtain feature vectors and  RAA is performed, as discussed in Sec \ref{sec:raa}.

% The second is a \emph{fine-tuned} attacker model, where the model is optimized on a training set of the target task. This requires a more sophisticated attacker, with the resources and computation to collect a dataset of labeled egocentric video and fine-tune the foundation model on this dataset. 
 
\noindent\textbf{Individual \& Situational Privacy} are formulated as a retrieval problem, with a suitable embedding model. Both query $\mathbf{x}$ and videos in ${\cal D}_\text{{test}}$ are mapped into the embedding to create feature vectors and those from ${\cal D}_{\text{test}}$ ranked by similarity to $\mathbf{x}$, using the {cosine similarity} metric. Capability \tikz[baseline=(char.base)]{\node[shape=circle,draw,inner sep=0.5pt] (char) {\textbf{1}}} is implemented by the embedding of the foundation model directly in a zero-shot manner. Capability \tikz[baseline=(char.base)]{\node[shape=circle,draw,inner sep=0.5pt] (char) {\textbf{2}}}: the embedding is fine-tuned on ${\cal D}_{\text{train}}$, as discussed in Sec \ref{sec:embed}.  The capability \tikz[baseline=(char.base)]{\node[shape=circle,draw,inner sep=0.5pt] (char) {\textbf{3}}} is only for demographic privacy and is thus omitted here.

\section{Retrieval-Augmented Attack}
\label{sec:raa}

We present a deeper dive into ego-to-exo retrieval under a novel \emph{retrieval-augmented} attack, to highlight its potential to boost the efficacy of classification-based attack models.

\begin{figure}[t]
    \centering
    \includegraphics[width=0.9\linewidth]{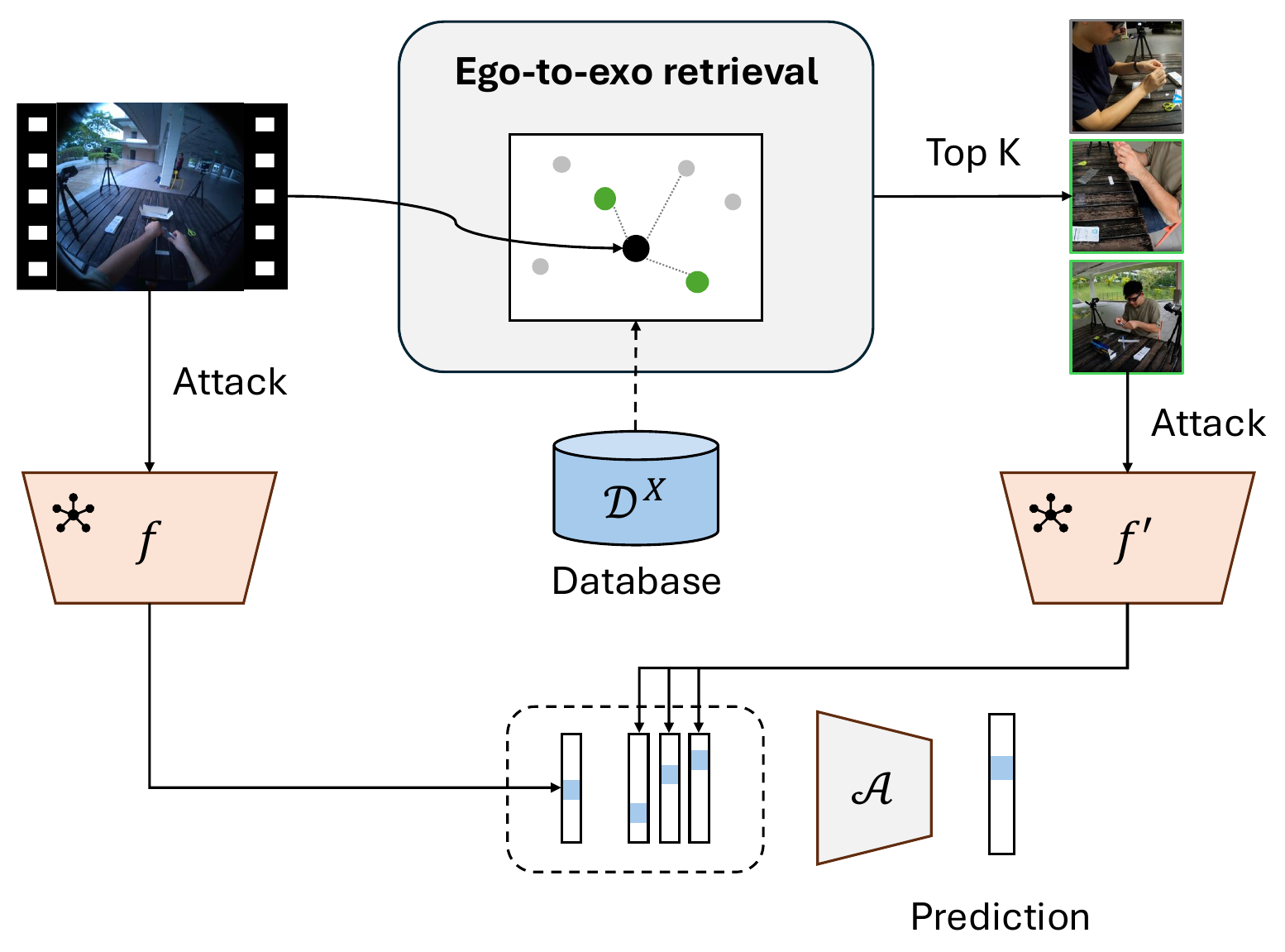}
    \caption{\textbf{Retrieval-Augmented Privacy Attacks.}}
    \label{fig:raa}
\end{figure}

\subsection{Ego-exo Embedding}
\label{sec:embed}
To perform ego-to-exo retrieval, a joint embedding space of ego and exo video clips is required. We follow recent progress on cross-modal metric learning~\cite{morgado2021audio, radford2021learning} and perform the ego-to-exo retrieval with an embedding learned by \emph{contrastive learning}~\cite{oord2018representation}.
A pair of egocentric $\mathbf{x}_i^E$ and exocentric $\mathbf{x}_i^X$ examples  is mapped into a pair of feature vectors $(\mathbf{z}_i^E, \mathbf{z}_i^X)$ using a joint embedding $(\mathbf{z}_i^E, \mathbf{z}_i^X) = (g(\mathbf{x}_i^E), g'(\mathbf{x}_i^X))$ where the mappings $g, g'$ are learned with a contrastive loss function. This uses ego-exo video pairs from the same person (demographic or individual privacy) or space-time (situational) as positive pairs. 
%If 1-to-1 mapping exists between egocentric and exocentric examples $(\mathbf{x}_i^E, \mathbf{x}_i^X)$, this can be achieved with the joint embedding s$(\mathbf{z}_i^E, \mathbf{z}_i^X) = (g(\mathbf{x}_i^E), g'(\mathbf{x}_i^X))$, where the mappings $g, g'$ are learned by optimizing the {InfoNCE} loss~\cite{oord2018representation}
%\begin{equation}
%    L(g, g') = -\sum_{i=1}^{N} \log \frac{\exp(\langle \mathbf{z}_i^E, \mathbf{z}_i^X \rangle / \tau)}{\sum_{j \neq i} \exp(\langle \mathbf{z}_i^E, \mathbf{z}_j^X \rangle / \tau)},
%    \label{eqn:infonce}
%\end{equation}
%where $\tau$ is a temperature hyperparameter. 

In general, several exocentric samples are associated with a single egocentric sample, either because the exocentric video is collected from multiple viewpoints or by definition of the retrieval task. For example, in individual privacy attacks all exocentric videos of the same camera wearer are considered successful retrievals, independently of whether they were shot at the same location or time. To account for this, we formulate the learning of the embedding as \emph{supervised contrastive learning} (SupCon)~\cite{khosla2020supervised}. This is a relaxed version of contrastive learning that distributes the loss evenly over all positive pairs
\begin{equation}
    \resizebox{\linewidth}{!}{
       $ \displaystyle L(g, g') = -\sum_{i=1}^{N} \frac{1}{\lvert P(i) \rvert} \sum_{k \in P(i)} \log \frac{\exp(\langle \mathbf{z}_i^E, \mathbf{z}_k^X \rangle / \tau)}{\sum_{j \in N(i)} \exp(\langle \mathbf{z}_i^E, \mathbf{z}_j^X \rangle/ \tau)},$
    }
    \label{eqn:supcon}
\end{equation}
where $P(i)$ is the set of exocentric feature vectors that are positive pairs of $\mathbf{z}_i^E$ and $N(i)$ a set of negative pairs. SupCon allows the unification of privacy types, individual and situational, simply by varying the definition of positive set $P(i)$. For individual privacy, $P(i)$ contains all exocentric examples $\mathbf{z}_i^X$ containing the camera wearer of $\mathbf{z}_i^E$. For situational privacy, $P(i)$ is restricted to the single exocentric video clip (single \emph{take} in Ego-Exo4D) recorded in sync with $\mathbf{x}_i^E$. In both cases, the negative set $N(i)$ is formed by all other exocentric examples in the same minibatch as well as cached from past iterations of training.

\subsection{Retrieval as Augmentation}
\label{sec:ra}
Egocentric video inherently offers greater privacy protection for the subject compared to exocentric video, as faces and most of the body are obscured. However, if an adversary has access to the identity mapping between egocentric and exocentric videos, they can easily infer private information from the exocentric footage. We further notice that the ego-to-exo retrieval attack model as discussed in Section \ref{attack_model} performs this task exactly. Motivated by this, we propose \emph{Retrieval Augmented Attack} ({RAA}) by exploiting an additional ego-to-exo retrieval model to retrieve exocentric videos for augmented prediction. 

% Specifically, given access to a large pool of \emph{third-person view} videos (e.g., uploaded by other users or from surveillance footage), one can obtain a set of videos likely showing the same camera wearer by applying ego-to-exo retrieval. Then the attack can perform privacy attacks on these exocentric videos using foundation models.

Formally, {RAA} is a two-stage privacy attack under the ``retrieve, then predict'' methodology, as illustrated in \cref{fig:raa}. {RAA} assumes the availability of an external pool of \emph{exocentric} data $\mathcal{D}^X$, which includes the individual behind the egocentric video. Given an egocentric query example $\mathbf{x}^E$, the attacker first uses an \emph{ego-to-exo retrieval} module $g$ to rank all examples $\mathbf{x}'_i \in \mathcal{D}^X$ by their similarity to $\mathbf{x}^E$ in the embedding space $s_{g, g'}(\mathbf{x}^E, \mathbf{x}'_i) = \langle g(\mathbf{x}^E), g'(\mathbf{x}'_i) \rangle$; a support set $\{\mathbf{x}_{1:M}^X\} \subset \mathcal{D}^X$ is then formed by the top-$M$ most similar examples.
The final output of RAA is the aggregation of the direct egocentric attack $f(\mathbf{x}^E)$ and the exocentric attacks on the retrieved examples $\{f'(\mathbf{x}_i^X)\}_{i=1}^M$:
\begin{equation}
    f^\textrm{RAA}(\mathbf{x}^E, \{\mathbf{x}_{1:M}^X\}) = \mathcal{A}\left(f(\mathbf{x}^E), f'(\mathbf{x}_1^X), \dots, f'(\mathbf{x}_M^X)\right)
    \label{eqn:raa}
\end{equation}
where $f, f'$ are classification-based privacy attacks, such as gender predictors, on egocentric and exocentric inputs,\footnote{One can use the same attack model for both views ($f = f'$).} and $\mathcal{A}$ is an aggregation function that can be as simple as majority voting (hard voting) or weighted pooling (soft voting). By employing the simple voting ensemble, RAA without bells and whistles demonstrates significant effectiveness, improving the attack rate by a large margin.

\newcommand{\gray}[0]{\cellcolor[HTML]{EFEFEF}}

\begin{table*}[t]
    \centering
\resizebox{0.95\linewidth}{!}{%
\scriptsize
\begin{tabular}{@{}lccc|cccr|cccr|cccr@{}}
\toprule
& \textbf{OOD} & \multicolumn{2}{c|}{\textbf{Capability}} & \multicolumn{4}{c|}{\textbf{Gender}} & \multicolumn{4}{c|}{\textbf{Race}} & \multicolumn{4}{c}{\textbf{Age}} \\
% \multicolumn{2}{c|}{\multirow{-2}{*}{\textit{\begin{tabular}[c]{@{}c@{}}Ego-Exo4D\\ (ID)\end{tabular}}}} 

& (Charades-Ego) & \tikz[baseline=(char.base)]{\node[shape=circle,draw,inner sep=0.5pt] (char) {\textbf{1}}} & \tikz[baseline=(char.base)]{\node[shape=circle,draw,inner sep=0.5pt] (char) {\textbf{2}}}
& 
{\color[HTML]{9B9B9B} Exo} & Ego & RAA (+ \tikz[baseline=(char.base)]{\node[shape=circle,draw,inner sep=0.5pt] (char) {\textbf{3}}}) & \multicolumn{1}{c|}{$\Delta$} 
& {\color[HTML]{9B9B9B} Exo} & Ego & RAA (+ \tikz[baseline=(char.base)]{\node[shape=circle,draw,inner sep=0.5pt] (char) {\textbf{3}}}) & \multicolumn{1}{c|}{$\Delta$} 
& {\color[HTML]{9B9B9B} Exo} & Ego & RAA (+ \tikz[baseline=(char.base)]{\node[shape=circle,draw,inner sep=0.5pt] (char) {\textbf{3}}}) & \multicolumn{1}{c}{$\Delta$} \\ 
\midrule
\rowcolor[HTML]{EFEFEF} 
Random Chance & & & & & {\cellcolor[HTML]{EFEFEF}50.00} & & \multicolumn{1}{c|}{\cellcolor[HTML]{EFEFEF}-} & & {\cellcolor[HTML]{EFEFEF}33.33} & & \multicolumn{1}{c|}{\cellcolor[HTML]{EFEFEF}-} & & {\cellcolor[HTML]{EFEFEF}33.33}  & & \multicolumn{1}{c}{\cellcolor[HTML]{EFEFEF}-} \\ 
\rowcolor[HTML]{EFEFEF} 
Prior & & & & & {\cellcolor[HTML]{EFEFEF}60.74} & & \multicolumn{1}{c|}{\cellcolor[HTML]{EFEFEF}-} & & {\cellcolor[HTML]{EFEFEF}54.17	}  & & \multicolumn{1}{c|}{\cellcolor[HTML]{EFEFEF}-} & & {\cellcolor[HTML]{EFEFEF}79.48}  & & \multicolumn{1}{c}{\cellcolor[HTML]{EFEFEF}-} \\ 

\midrule
\multirow{1}{*}{Hand-based} & \xmark & \multicolumn{2}{c|}{N/A} & {\color[HTML]{9B9B9B} -} & {45.33} & {-} & \multicolumn{1}{c}{-} & \multicolumn{1}{c}{-} & {-} & {-} & \multicolumn{1}{c}{-} & {\color[HTML]{9B9B9B} -} & {65.30} & {-} & \multicolumn{1}{c}{-}  \\
\multirow{1}{*}{Face-based} & \xmark & \multicolumn{2}{c|}{N/A} & {\color[HTML]{9B9B9B} 70.98} & {-} & {-} & \multicolumn{1}{c}{-} & {\color[HTML]{9B9B9B} -} & {-} & {-} & \multicolumn{1}{c}{-} & {\color[HTML]{9B9B9B} 69.57} & {-} & {-} & \multicolumn{1}{c}{-}  \\
\midrule
% \multicolumn{14}{c}{\textit{ID (EgoExo-4D testset)}} \\ 
% \midrule
\multirow{4}{*}{CLIP$_{\text{H/14}}$} & \xmark & \cmark & \xmark & {\color[HTML]{9B9B9B} 78.64} & 57.89 & 67.35 & {\color[HTML]{009901} 9.46} & {\color[HTML]{9B9B9B} 60.04} & 45.21 & 60.98 & {\color[HTML]{009901} 15.77} & {\color[HTML]{9B9B9B} 73.51} & 72.02 & 76.23 & {\color[HTML]{009901} 4.21} \\
 & \xmark & \xmark & \cmark & {\color[HTML]{9B9B9B} 88.33} & {  68.87} & 76.98 & \multicolumn{1}{r|}{{\color[HTML]{009901} 8.11}} & {\color[HTML]{9B9B9B} 73.93} & {  70.92} & 71.92 & \multicolumn{1}{r|}{{\color[HTML]{009901} 1.00}} & {\color[HTML]{9B9B9B} 77.15} & 79.73 & 79.73 & 0.00 \\
  & \cmark& \cmark & \xmark & {\color[HTML]{9B9B9B} 89.80} & 70.00 & { 77.31} & \multicolumn{1}{r|}{{\color[HTML]{009901} 7.31}} & {\color[HTML]{9B9B9B} 60.14} & 46.09 & 59.42 & \multicolumn{1}{r|}{{\color[HTML]{009901} 13.33}} & {\color[HTML]{9B9B9B} 48.02} & 20.75 & 26.42 & {\color[HTML]{009901} 5.67} \\
 & \cmark& \xmark & \cmark &  {\color[HTML]{9B9B9B} 75.12} & { 54.70} & {69.65} & \multicolumn{1}{r|}{{\color[HTML]{009901} 14.95}} & {\color[HTML]{9B9B9B} 85.01} & {63.68} & {74.09} & \multicolumn{1}{r|}{{\color[HTML]{009901} 10.41}} & {\color[HTML]{9B9B9B}29.90 } & 29.70 & 29.92 & {\color[HTML]{009901} 0.22} \\
\midrule
\multirow{3}{*}{EgoVLP v2} & \xmark & \cmark & \xmark & {\color[HTML]{9B9B9B} 76.97} & {63.18} & 67.11 & \multicolumn{1}{r|}{{\color[HTML]{009901} 3.93}} & {\color[HTML]{9B9B9B} 64.85} & {57.14} & {64.29} & \multicolumn{1}{r|}{{\color[HTML]{009901} 7.15}} & {\color[HTML]{9B9B9B} 52.25} & 47.88 & {49.67} & {\color[HTML]{009901} 1.79} \\

& \xmark & \xmark & \cmark & {\color[HTML]{9B9B9B} 84.85} & {71.81} & { 77.88} & \multicolumn{1}{r|}{{\color[HTML]{009901} 6.07}} & {\color[HTML]{9B9B9B} 71.46} & {72.01} & {75.57} & \multicolumn{1}{r|}{{\color[HTML]{009901} 3.56}} & {\color[HTML]{9B9B9B} 77.11} & {80.72} & {81.88} & {\color[HTML]{009901} 1.16}  \\
 & \cmark & \xmark & \cmark & {\color[HTML]{9B9B9B} 77.09} & {56.27} & 68.38 & \multicolumn{1}{r|}{{\color[HTML]{009901} 12.11}} & {\color[HTML]{9B9B9B} 78.25} & {  62.82} & {  69.14} & \multicolumn{1}{r|}{{\color[HTML]{009901} 6.32}} & {\color[HTML]{9B9B9B} 30.57} & 29.70 & {30.96} & {\color[HTML]{009901} 1.26} \\
\midrule
\multirow{2}{*}{VideoMAE$_{\text{B/14}}$} & \xmark & \xmark & \cmark & {\color[HTML]{9B9B9B} 72.42} & 63.69 & 70.65 & \multicolumn{1}{r|}{{\color[HTML]{009901} 6.96}} & {\color[HTML]{9B9B9B} 75.16} & 66.73 & {  73.49} & \multicolumn{1}{r|}{{\color[HTML]{009901} 6.76}} & {\color[HTML]{9B9B9B} 78.21} & 79.73 & {  81.70} & {\color[HTML]{009901} 1.97} \\
 & \cmark & \xmark & \cmark &  {\color[HTML]{9B9B9B} 67.97} & 42.09 & 55.40 & \multicolumn{1}{r|}{{\color[HTML]{009901} 13.31}} & {\color[HTML]{9B9B9B} 72.08} & 46.50 & 57.42 & \multicolumn{1}{r|}{{\color[HTML]{009901} 10.92}} & {\color[HTML]{9B9B9B} 30.57} & 29.70 & {  30.33} & {\color[HTML]{009901} 0.63}  \\

\multirow{2}{*}{VideoMAE$_{\text{L/14}}$} & \xmark & \xmark & \cmark & {\color[HTML]{9B9B9B} 87.14} & 63.87 & {78.95} & \multicolumn{1}{r|}{{\color[HTML]{009901} 16.08}} & {\color[HTML]{9B9B9B} 74.36} & 70.10 & 72.65 & \multicolumn{1}{r|}{{\color[HTML]{009901} 2.55}} & {\color[HTML]{9B9B9B} 77.15} & 79.73 & 79.73 & 0.00 \\
& \cmark & \xmark & \cmark & {\color[HTML]{9B9B9B} 80.67} & 54.63 & {  68.44} & \multicolumn{1}{r|}{{\color[HTML]{009901} 13.81}} & {\color[HTML]{9B9B9B} 72.37} & 46.02 & 57.42 & \multicolumn{1}{r|}{{\color[HTML]{009901} 11.40}} & {\color[HTML]{9B9B9B} 29.90} & 29.70 & 29.92 & {\color[HTML]{009901} 0.22} \\

\midrule
\multirow{2}{*}{$\text{LLaVA-1.5}_{\text{7B}}$} & \xmark & \cmark & \xmark & {\color[HTML]{9B9B9B} 91.52} & 66.90 & 77.16 & \multicolumn{1}{r|}{{\color[HTML]{009901} 10.26}} & {\color[HTML]{9B9B9B} 60.06} & 57.34 & 57.52 & \multicolumn{1}{r|}{{\color[HTML]{009901} 0.18}} & {\color[HTML]{9B9B9B} 79.29} & {  79.46} & {  79.55} & {\color[HTML]{009901} 0.09} \\
 & \cmark & \cmark & \xmark & {\color[HTML]{9B9B9B} 90.42} & {  71.59} & 75.60 & \multicolumn{1}{r|}{{\color[HTML]{009901} 4.01}} & {\color[HTML]{9B9B9B} 71.10} & 48.95 & 59.32 & \multicolumn{1}{r|}{{\color[HTML]{009901} 10.37}} & {\color[HTML]{9B9B9B} 50.33} & 35.07 & 47.26 & {\color[HTML]{009901} 12.19}  \\

\multirow{2}{*}{$\text{LLaVA-1.5}_{\text{13B}}$} & \xmark & \cmark & \xmark & {\color[HTML]{9B9B9B} 90.37} & 65.45 & 78.55 & \multicolumn{1}{r|}{{\color[HTML]{009901} 13.10}} & {\color[HTML]{9B9B9B} 66.64} & { 62.81} & {69.33} & \multicolumn{1}{r|}{{\color[HTML]{009901} 6.52}} & {\color[HTML]{9B9B9B} 78.55} & 69.33 & 72.56 & {\color[HTML]{009901} 3.23} \\
 & \cmark & \cmark & \xmark &   {\color[HTML]{9B9B9B} 88.38} & 62.37 & 72.61 & \multicolumn{1}{r|}{{\color[HTML]{009901} 10.24}} & {\color[HTML]{9B9B9B} 70.48} & 46.42 & 59.32 & \multicolumn{1}{r|}{{\color[HTML]{009901} 12.90}} & {\color[HTML]{9B9B9B} 51.56} & {  37.44} & {  47.56} & {\color[HTML]{009901} 10.12} \\
\midrule
\multirow{2}{*}{$\text{VideoLLaMA2}_{\text{7B}}$} & \xmark & \cmark & \xmark & {\color[HTML]{9B9B9B} 90.96} & {73.15} & {79.48} & \multicolumn{1}{r|}{{\color[HTML]{009901} 6.33}} & {\color[HTML]{9B9B9B} 71.53} & 53.97 & { 69.10} & \multicolumn{1}{r|}{{\color[HTML]{009901} 15.13}} & {\color[HTML]{9B9B9B} 52.99} & 47.08 & 56.14 & {\color[HTML]{009901} 9.06} \\
 & \cmark & \cmark & \xmark & {\color[HTML]{9B9B9B} 90.69} & 71.31 & 76.16 & \multicolumn{1}{r|}{{\color[HTML]{009901} 4.85}} & {\color[HTML]{9B9B9B} 75.56} & { 62.39} & { 68.48} & \multicolumn{1}{r|}{{\color[HTML]{009901} 6.09}} & {\color[HTML]{9B9B9B} 64.99} & {57.11} & {59.03} & {\color[HTML]{009901} 1.92} \\

\multirow{2}{*}{$\text{VideoLLaMA2}_{\text{72B}}$} & \xmark & \cmark & \xmark & {\color[HTML]{9B9B9B} 91.59} & {  70.03} & 78.41 & \multicolumn{1}{r|}{{\color[HTML]{009901} 8.38}} & {\color[HTML]{9B9B9B} 69.25} & {65.36} & 67.82 & \multicolumn{1}{r|}{{\color[HTML]{009901} 2.46}} & {\color[HTML]{9B9B9B} 82.46} & {79.64} & {81.26} & {\color[HTML]{009901} 1.62} \\
 & \cmark & \cmark & \xmark & {\color[HTML]{9B9B9B} 92.25} & {73.74} & {77.89} & \multicolumn{1}{r|}{{\color[HTML]{009901} 4.15}} & {\color[HTML]{9B9B9B} 76.71} & {66.58} & {69.04} & \multicolumn{1}{r|}{{\color[HTML]{009901} 2.46}} & {\color[HTML]{9B9B9B} 55.88} & 32.93 & 45.23 & {\color[HTML]{009901} 12.30} \\

\bottomrule
\end{tabular}%
}
    \caption{Results on \textbf{Demographic Privacy}. Accuracy is calculated on a \emph{per-video} basis. $\Delta$ indicates the accuracy increase brought by RAA (\tikz[baseline=(char.base)]{\node[shape=circle,draw,inner sep=0.5pt] (char) {\textbf{3}}}) over \tikz[baseline=(char.base)]{\node[shape=circle,draw,inner sep=0.5pt] (char) {\textbf{1}}} / \tikz[baseline=(char.base)]{\node[shape=circle,draw,inner sep=0.5pt] (char) {\textbf{2}}}.}
    \label{tab:results_demographic}
\end{table*}

\section{Results}
\label{sec:results}

\subsection{Experimental Setup}

\textbf{Objectives.} We begin with a set of research questions and objectives of the experiments:
% \vspace{-2mm}
\begin{itemize}[nosep, leftmargin=*]
    \item Are egocentric videos a threat to the privacy of the camera wearer?
    \item To what extent do egocentric videos expose private information with different capabilities of the threat model?
    \item How effective is {RAA} in enhancing privacy attacks?
    \item What factors contribute to privacy vulnerabilities in egocentric videos?
    \item Do privacy attacks remain effective for out-of-distribution samples?
\end{itemize}

\textbf{Dataset.}
All experiments are performed on the \ours{} benchmark
%, built with videos from the Ego-Exo4D~\cite{grauman2024ego} and Charades-Ego~\cite{sigurdsson2018charades} datasets, as 
discussed in Section \ref{sec:benchmark}.

\textbf{Models \& Baselines.}
We consider a variety of models for launching the privacy attack, ranging from generalist vision-language models like CLIP~\cite{radford2021learning,fang2023data} to video-centric models such as VideoMAE \cite{tong2022videomae} and EgoVLPv2~\cite{pramanick2023egovlpv2} pre-trained on egocentric data, and large multimodel models (LMMs), such as LLaVA-1.5~\cite{liu2024visual} and VideoLLaMA2~\cite{cheng2024videollama}.

For exocentric demographic attacks, we also consider a straightforward face-based baseline, i.e.\ run face detection and demographic classification. Given the discovery that hand-based biometrics can be leveraged for inferring demographics such as gender and race~\cite{matkowski2019palmprint,matkowski2020gender}, we also employed a hand-based demographics classifier as a baseline for egocentric demographic attacks.   

\textbf{Training.} We add to the top of the foundation models with one layer of MLP for classification (demographic privacy) and use its representation layer for retrieval (individual and situational privacy). All models are trained with 1$\times$A100 with a batch size of 8. We use a learning rate of 1e-5 and adopt the AdamW optimizer with cosine learning rate decay. The default number of frames for one video is 8.

\begin{table*}[t]
\centering
% \scriptsize
\tiny
\begin{tabular}{@{}lcc|cccccccc@{}}
\toprule
 &  &     &   \multicolumn{4}{c|}{\textit{Identity}} & \multicolumn{4}{c}{\textit{Situational}} \\ \midrule
 &  &        \multicolumn{1}{c|}{} & \multicolumn{2}{c|}{\textbf{Ego$\to$Ego}} & \multicolumn{2}{c|}{\textbf{Ego$\to$Exo}} & \multicolumn{2}{c|}{\textbf{Scene}} & \multicolumn{2}{c}{\textbf{Moment}} \\
  &    \multicolumn{1}{c}{\multirow{1}{*}{\tikz[baseline=(char.base)]{\node[shape=circle,draw,inner sep=0.5pt] (char) {\textbf{1}}}}} & \multicolumn{1}{c|}{\multirow{1}{*}{\tikz[baseline=(char.base)]{\node[shape=circle,draw,inner sep=0.5pt] (char) {\textbf{2}}}}}  & HR@1 & \multicolumn{1}{c|}{HR@5} & HR@1 & \multicolumn{1}{c|}{HR@5} & HR@1 & HR@5 & HR@1 & HR@5 \\ \midrule
\rowcolor[HTML]{EFEFEF} 
Random Chance    & \multicolumn{2}{c|}{\cellcolor[HTML]{EFEFEF}N/A} & 0.57 & \multicolumn{1}{c|}{\cellcolor[HTML]{EFEFEF}2.87} & 0.57 & \multicolumn{1}{c|}{2.87} & {\color[HTML]{9B9B9B} } & \multicolumn{1}{c|}{\cellcolor[HTML]{EFEFEF}} & 0.09 & 0.45 \\ 
% Prior  &      & \multicolumn{1}{c|}{\cellcolor[HTML]{EFEFEF}-} & 0.57 & \multicolumn{1}{c|}{\cellcolor[HTML]{EFEFEF}2.87} & 0.57 & \multicolumn{1}{c|}{2.87} & {\color[HTML]{9B9B9B} } & \multicolumn{1}{c|}{\cellcolor[HTML]{EFEFEF}} & 0.09 & 0.45 \\ 
\midrule
\multicolumn{11}{c}{\textit{ID (EgoExo4D testset)}}\\
\midrule
 \multirow{2}{*}{CLIP$_{\text{H/14}}$}   &   \cmark  &  \xmark  & 0.92 & \multicolumn{1}{c|}{1.10} & 0.89 & \multicolumn{1}{c|}{1.07} & {24.98} & \multicolumn{1}{c|}{29.07} & 1.78 & 7.94 \\
   &     \xmark   & \cmark & {  79.37} & \multicolumn{1}{c|}{{  96.97}} & {  49.69} & \multicolumn{1}{c|}{{  63.51}} & \textbf{89.21} & \multicolumn{1}{c|}{\textbf{89.56}} & {  13.21} & {  39.57} \\
\midrule
 \multirow{2}{*}{EgoVLP v2}&   \cmark &  \xmark  & 4.85 & \multicolumn{1}{c|}{8.31} & 7.31 & \multicolumn{1}{c|}{18.38} & {28.64} & \multicolumn{1}{c|}{28.88} & 1.96 & 7.94 \\ 
 &    \xmark   &  \cmark  & \textbf{81.25} & \multicolumn{1}{c|}{\textbf{97.34}} & \textbf{50.31} & \multicolumn{1}{c|}{\textbf{66.82}} & {  84.92} & \multicolumn{1}{c|}{{  87.96}} & \textbf{15.43} & \textbf{43.00} \\
 \midrule
 \multirow{2}{*}{VideoMAE$_{\text{B}}$}    & \cmark  &  \xmark  & 0.49 & \multicolumn{1}{c|}{1.35} & 0.68 & \multicolumn{1}{c|}{1.02} & {14.32} & \multicolumn{1}{c|}{16.37} & 0.09 & 0.71 \\ 
    &    \xmark   & \cmark  & 63.47 & \multicolumn{1}{c|}{84.96} & 24.84 & \multicolumn{1}{c|}{36.09} & 69.09 & \multicolumn{1}{c|}{69.44} & 10.52 & 33.49 \\ 
  \multirow{2}{*}{VideoMAE$_{\text{L}}$} &  \cmark  &  \xmark   & 0.88 & \multicolumn{1}{c|}{1.74} & 0.93 & \multicolumn{1}{c|}{1.07} & {13.60} & \multicolumn{1}{c|}{15.98} & 0.00 & 0.45 \\ 
   &  \xmark  & \cmark  & 62.91 & \multicolumn{1}{c|}{79.38} & 24.29 & \multicolumn{1}{c|}{38.21} & {70.32} & \multicolumn{1}{c|}{71.49} & 9.42 & 32.29 \\ 
 \midrule
\multicolumn{11}{c}{\textit{OOD (Charades-Ego testset)}} \\ 
\midrule
 \multirow{2}{*}{CLIP$_{\text{H/14}}$}   & \cmark  &  \xmark  & 0.59 & \multicolumn{1}{c|}{1.04} & 0.90 & \multicolumn{1}{c|}{1.89} & {-} & \multicolumn{1}{c|}{-} & 1.49 & 6.53 \\
   &    \xmark   & \cmark & {  71.42} & \multicolumn{1}{c|}{\textbf{93.57}} & {  39.50} & \multicolumn{1}{c|}{{  58.01}} & {-} & \multicolumn{1}{c|}{-} & {  11.55} & \textbf{37.90} \\
\midrule
 \multirow{2}{*}{EgoVLP v2}&   \cmark &   \xmark   & 5.03 & \multicolumn{1}{c|}{9.90} & 6.74 & \multicolumn{1}{c|}{17.44} & {-} & \multicolumn{1}{c|}{-} & 1.77 & 6.53 \\ 
 &  \xmark  & \cmark  & \textbf{72.48} & \multicolumn{1}{c|}{{  85.69}} & \textbf{45.33} & \multicolumn{1}{c|}{\textbf{62.09}} & {-} & \multicolumn{1}{c|}{-} & \textbf{12.74} & {  37.74} \\
 \midrule
 \multirow{2}{*}{VideoMAE$_{\text{L}}$}  & \cmark &   \xmark  & 0.69 & \multicolumn{1}{c|}{1.49} & 0.83 & \multicolumn{1}{c|}{1.95} & {-} & \multicolumn{1}{c|}{-} & 0.42 & 0.99 \\ 
    &  \xmark  & \cmark  & 58.02 & \multicolumn{1}{c|}{81.83} & 22.80 & \multicolumn{1}{c|}{36.94} & {-} & \multicolumn{1}{c|}{-} & 9.44 & 30.08 \\ 
    \multirow{2}{*}{VideoMAE$_{\text{L}}$} & \cmark   &  \xmark  & 0.57 & \multicolumn{1}{c|}{1.58} & 1.04 & \multicolumn{1}{c|}{2.38} & {-} & \multicolumn{1}{c|}{-} & 0.48 & 1.16 \\ 
   &  \xmark & \cmark  & 60.09 & \multicolumn{1}{c|}{80.33} & 23.57 & \multicolumn{1}{c|}{39.32} & {-} & \multicolumn{1}{c|}{-} & 9.09 & 29.57 \\ 
 \bottomrule
\end{tabular}%
\vspace{-0.1 in}
\caption{Results on \textbf{Identity} and \textbf{Situational Privacy}. The hit rate is calculated on a per-video basis. Scene retrieval results are omitted for \textit{OOD (Charades-Ego test set)} due to the absence of ground-truth labels in \textit{Charades-Ego} dataset.}
\label{tab:results2}
\end{table*}

\subsection{Main Results} 
\label{main_result}
\textbf{Are egocentric videos a threat to the privacy of the camera wearer?} We answer this by comparing different models with chance-level (lower bound) and exocentric performance (upper bound). As per Tables~\ref{tab:results_demographic} and \ref{tab:results2}, we can clearly observe that 1) despite some lower than exocentric performance, all attack models in Tables~\ref{tab:results_demographic} are higher than random chance by a large margin (more than 15\%) for both Demographic, Identity and Situational Privacy; 2) except for zero-shot models, all fine-tuned models in Table \ref{tab:results2} achieve significantly higher results compared to chance-level performance. The unsatisfactory performance of the zero-shot retrieval model is attributed to the fact that some of these models have not been trained on egocentric videos before, and hence fail to construct a meaningful ego-view representation. These results suggest that the risk of privacy leakage is a significant concern in egocentric vision.

\textbf{To what extent do egocentric videos expose private information under different capabilities of the threat model?} We evaluate the attack performance under a threat model with different capabilities outlined in \cref{threat_model}. First, using zero-shot foundation models (\tikz[baseline=(char.base)]{
            \node[shape=circle,draw,inner sep=0.5pt] (char) {\textbf{1}};}), we observe a really high \emph{demographic} attack accuracy in Table \ref{tab:results_demographic}, as illustrated by the highest 73.15\%, 65.36\% and 79.64\% for gender, race and age respectively. This leads to the conclusion that even with minimum capabilities, the adversary can still perform a successful attack with up to 80\% success rate. However, zero-shot models perform significantly worse on \emph{situational} and \emph{identity} attacks (\cref{tab:results2}), leaving these two privacy protected against capability \tikz[baseline=(char.base)]{
            \node[shape=circle,draw,inner sep=0.5pt] (char) {\textbf{1}};}.

            When equipped with a training dataset (\tikz[baseline=(char.base)]{
            \node[shape=circle,draw,inner sep=0.5pt] (char) {\textbf{2}};}), race and age results can be further improved to 72.01\% and 80.72\%, and retrieval-based attacks reach the highest of 81.2\%, 50.31\%, 89.21\% and 15.43\% top-1 hit rate on ego-to-ego, ego-to-exo identity, scene and moment retrieval tasks respectively. This suggests that, with access to some training data, an adversary could further extract more private information about the camera wearer from egocentric videos, thereby posing an even greater threat to privacy.

\textbf{Effectiveness of RAA.} With the additional capability \tikz[baseline=(char.base)]{
            \node[shape=circle,draw,inner sep=0.5pt] (char) {\textbf{3}};}, adversary is now able to perform the RAA attack. We demonstrate the delta after and before applying the RAA in Table \ref{tab:results_demographic}. We can see a consistent improvement over all the models across all three tasks, with some even surpassing the exocentric baseline (e.g. EgoVLP v2). The most significant improvement is observed with the VideoMAE model on the gender classification task, achieving an increase in accuracy of over 16\%. This result has demonstrated the effectiveness of RAA in most scenarios. We also observe some minimal improvement cases. These cases can be attributed to the small gap between egocentric and exocentric performance, leading to a minimal increase. We believe this is reasonable, as the performance on exocentric is generally seen as the upper bound of an egocentric privacy attack. 

            We also notice that, even when the exocentric performance is lower than egocentric, RAA still offers improvements in some cases. We derive a hypothesis that RAA does not need the retrieval model to select the correct identity necessary to improve, but rather the retrieval model will cluster and group identities of similar attributes (of same gender, age and race, etc). To validate such a hypothesis, we conduct an experiment to see whether the ego-to-exo model groups identities of similar gender, race and age together. Specifically, we test how many top-1 and top-5 retrieved identities are of the same gender, age and race, as shown in Table \ref{tab:ablation1}. We can see that these retrieval models group people with similar gender, age and race together at a chance of over 82\%, much higher than the chance it selects the correct identity (which is 50.31\%). As long as the retrieval selects the identities with the correct demographic attributes, RAA can be improve the demographic classification.

\begin{table}[h]
\small
% \begin{center}
% \begin{tabular}{lcc}
% \toprule
% Demographic & Top-1 & Top-5 \\ \midrule
% Gender      & 82.22 & 89.83 \\ \midrule
% Age         & 84.51 & 90.74 \\ \midrule
% race   & 82.95 & 87.53 \\ \bottomrule
% \end{tabular}
% \end{center}
\begin{center}
\scalebox{0.9}{
\begin{tabular}{cccccc}
\toprule
\multicolumn{2}{c}{Gender} & \multicolumn{2}{c}{Age} &\multicolumn{2}{c}{Race}\\
\midrule
 Top-1 & Top-5&  Top-1 & Top-5& Top-1 & Top-5\\ 
 \midrule
 82.22 & 89.83 & 84.51 & 90.74 & 82.95 & 87.53\\ 
 \bottomrule
\end{tabular}}
\end{center}
\vspace{-0.2in}
\caption{Exo-to-ego identity retrieval as a demographic classifier.}
\label{tab:ablation1}
\vspace{-0.2in}
\end{table}

\begin{figure}
\centering
\includegraphics[width=0.3\textwidth]{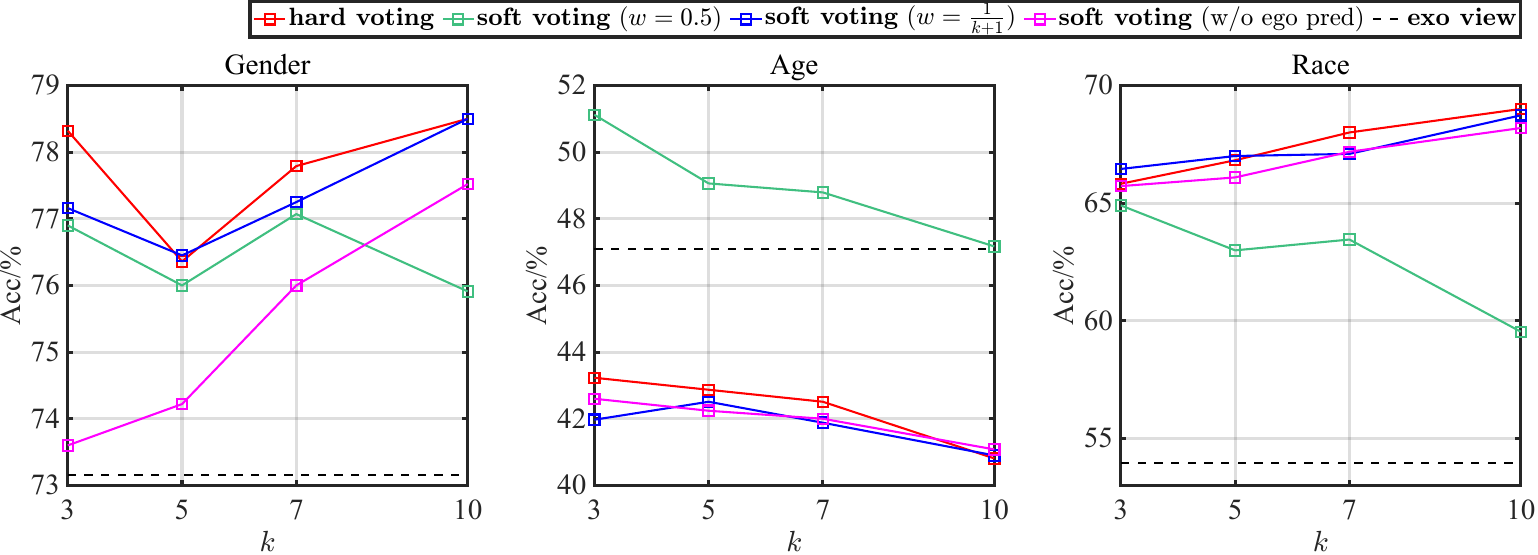}
\vspace{-0.2in}
\caption{Performance of Retrieval Augmented Attack versus $k$.}
\label{figure_topk}
\end{figure}

\begin{table}
\centering
\resizebox{.4\textwidth}{!}{
\small
\begin{tabular}{llccc}
\toprule
                                &              & Gender & Race & Age   \\ \midrule
\multicolumn{2}{l}{$\text{VideoLLaMA2}_{\text{7B}}$}            & 73.15                       & 53.97                          & \underline{47.08 }\\ \midrule
\multicolumn{2}{l}{- w/ hard voting}            & \textbf{78.32}                       & \underline{65.82  }                        & 43.23 \\ \midrule
\multicolumn{1}{l|}{\multirow{2}{*}{- w/ soft voting}} & $w=0.5$        & 76.90                        & 64.90                          & \textbf{51.12} \\ 
        \multicolumn{1}{c|}{}                         & $w=1 / (k + 1)$ & \underline{77.16}                       & \textbf{66.45 }                         & 41.97 \\\bottomrule

\end{tabular}}
\vspace{-0.1in}
\caption{\small Performance with different voting mechanisms for the Retrieval Augmented Attack. $w$ here refers to the weight over the egocentric prediction.}
\label{tab:voting}
\end{table}

\paragraph{Ablation study on voting parameters.} As discussed in Section \ref{sec:ra}, RAA retrieves the top $k$ exocentric views to augment the egocentric view for prediction. Given these $k$ exocentric predictions and one egocentric prediction, an ensemble method is required to effectively combine them into a final output. In this Section, we explore two ensemble strategies and conduct ablation studies on various hyperparameters. Hard voting, the simplest approach, involves voting on the predicted category and selecting the majority class. Given $k + 1$ predictions $f_1, \cdots, f_{k + 1}$, 
\vspace{-2mm}
$$\small \hat y
=
\arg\max_{c \in \mathcal{Y}}
\sum_{i=1}^{k+1}
\mathbbm{1}[f_i(x) = c].
$$
We also consider weighted soft voting, where we weighted sum the predicted probabilities from the $k+1$ views (softmax over logits)  and use the category with the highest aggregated probability as the final prediction.
\vspace{-2mm}
$$\small \hat y
=
\arg\max_{c \in \mathcal{Y}}
\sum_{i=1}^{k+1}
w_i f_i(x)
$$
where $w_i$ is the weight for prediction from view $i$
As shown in Table~\ref{tab:voting}, both hard and soft voting improve performance compared to the egocentric baselines. Hard voting generally yields better results for gender prediction, while soft voting consistently outperforms across all three demographic attributes. Therefore, we adopt soft voting as the default ensemble method.
We further ablate the effect of the choice of $w$ in the soft voting ensemble, as shown in Table~\ref{tab:voting}. Specifically, we compare two approaches: assigning evenly distributed weights ($w = \frac{1}{k + 1}$) and assigning a weight of 0.5 to the egocentric prediction ($w = 0.5$).
We also ablate the effect of the $k$ in top-$k$ retrieval in Figure \ref{figure_topk}, where $k=3$ leads to the optimal performance for Gender and Age. For Race, we observe that a larger $k=3$ leads to increasing performance.

\textbf{Can privacy attacks remain effective against out-of-distribution samples?} This question is practical, as privacy attacks often occur in real-world scenarios where in-distribution data is difficult to obtain. We use CharadesEgo as the OOD test set and evaluate all the attacker models described above, as presented in Table~\ref{tab:results_demographic} and Table~\ref{tab:results2}. We observe a consistent performance drop on the OOD data for all fine-tuned models, whereas the zero-shot foundation model maintains its original performance. This indicates a degree of overfitting during the fine-tuning stage and further underscores the privacy challenges inherent to egocentric videos: even with minimal attack capabilities (i.e., a zero-shot foundation model), an adversary can still launch effective attacks across varying data distributions.

\textbf{Capability }\tikz[baseline=(char.base)]{
        \node[shape=circle,draw,inner sep=0.5pt] (char) {\textbf{4}};} 
        As discussed in Section \ref{threat_model}, Capability \tikz[baseline=(char.base)]{
        \node[shape=circle,draw,inner sep=0.5pt] (char) {\textbf{4}};} further assumes the ability of adversary to ascertain whether two egocentric videos share the same identity, therefore enabling it to ensemble the predictions over all the videos and infer the demographic attributes of the identity more effectively. We repeat the demographic privacy attacks of Table~\ref{tab:results_demographic}, but assume the additional Capability \tikz[baseline=(char.base)]{
        \node[shape=circle,draw,inner sep=0.5pt] (char) {\textbf{4}};} of the adversary. We present the result in Appendix \ref{ability4} due to limited space. Equipped with Capability \tikz[baseline=(char.base)]{
        \node[shape=circle, draw, inner sep=0.5pt] (char) {\textbf{4}};}, despite an improved performance on Gender egocentric and all exocentric videos, the performance drops on the rest of the tasks, surprisingly.

\begin{figure}
\centering
\includegraphics[width=0.4\textwidth]{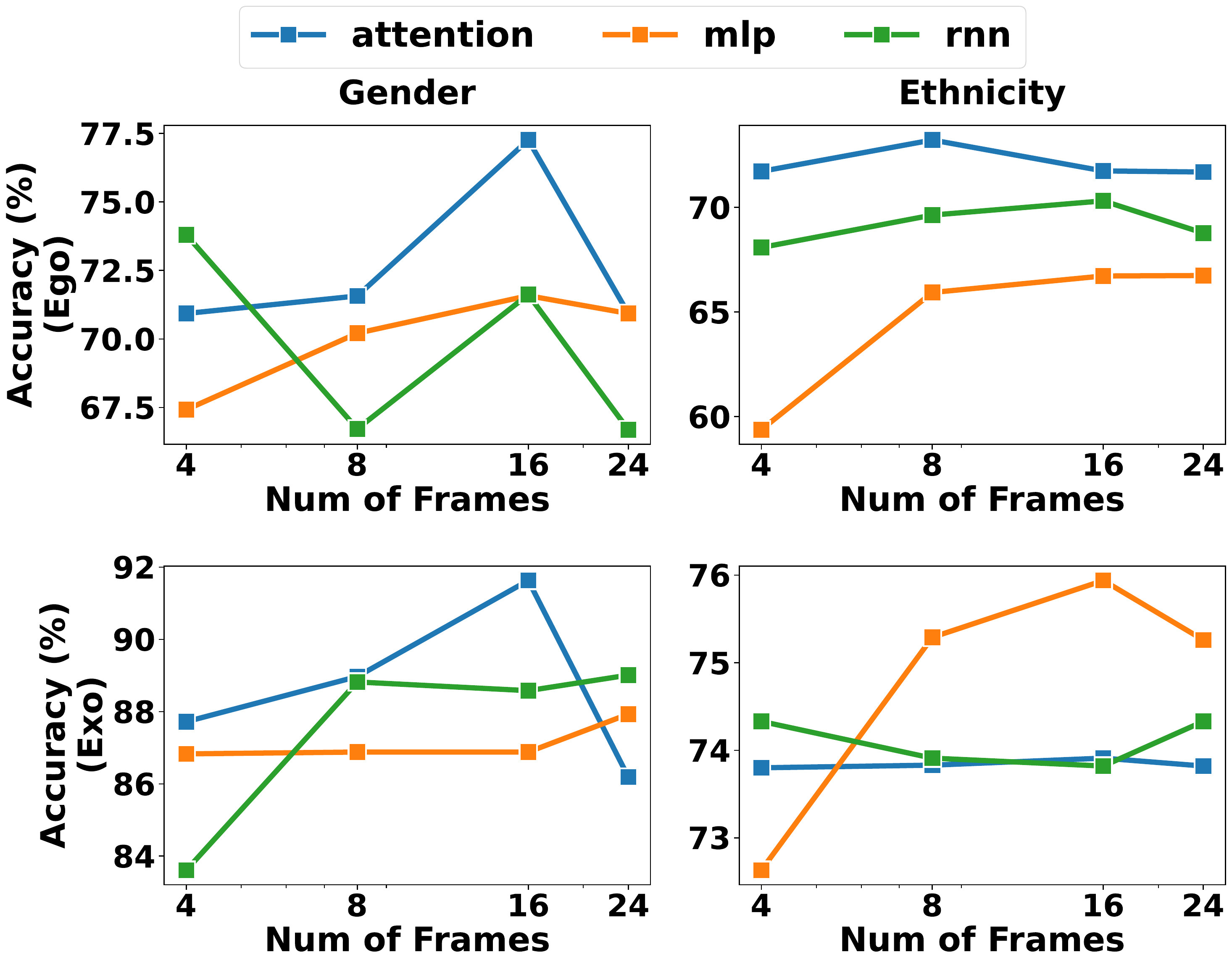}
\vspace{-0.1in}
\caption{Performance of CLIP model with MLP, RNN, and attention head on Demographic Privacy.}
\label{figure_temporal_attr}
\vspace{-0.08in}
\end{figure}

\textbf{What factors influence attacker models?}
A preliminary comparison in Table~\ref{tab:results_demographic} and Table~\ref{tab:results2} shows that EgoVLPv2 Fine-tuned consistently outperforms CLIP Fine-tuned, suggesting that temporal modeling aids adversaries in revealing private information. To investigate this effect, we evaluated models with MLP, Attention, and RNN layers atop the CLIP backbone, controlling for the number of parameters in each head. MLP layers map features to categories without temporal modeling, while Attention and RNN layers incorporate temporal information (temporal position embedding in Attention and recurrent nature of RNN). As shown in Figure~\ref{figure_temporal_attr}: (1) Increasing the number of frames improves performance ($4\Rightarrow8$), but saturates beyond 8 or 16 frames; (2) Temporal modeling (Attention or RNN) consistently outperforms MLP. This effect is more pronounced. These findings are further validated for Identity and Situational Privacy in \cref{app:addi_temporal_effect}.

\newcommand\cellwidthteaser{1.}

\begin{table}
\begin{center}
\setlength{\tabcolsep}{1pt} % Default value: 6pt
\renewcommand{\arraystretch}{1.2}
\scriptsize 
\begin{tabular}{cccccc}
\toprule
 \multicolumn{2}{c}{Age} &  \multicolumn{2}{c}{Gender} &  \multicolumn{2}{c}{Race}  
\tabularnewline
 Ego & Exo  & Ego & Exo  & Ego & Exo 
\tabularnewline
\midrule
{{\includegraphics[width=\cellwidthteaser cm, height=\cellwidthteaser cm]{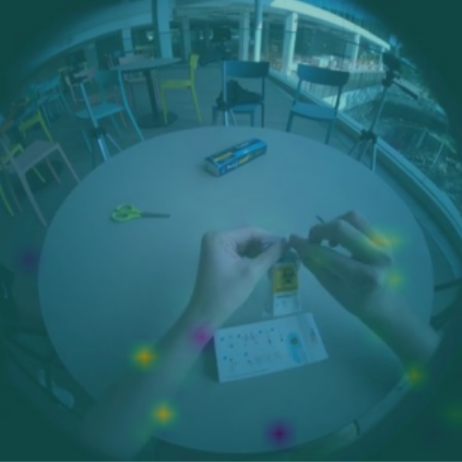}}} &
{{\includegraphics[width=\cellwidthteaser cm, height=\cellwidthteaser cm]{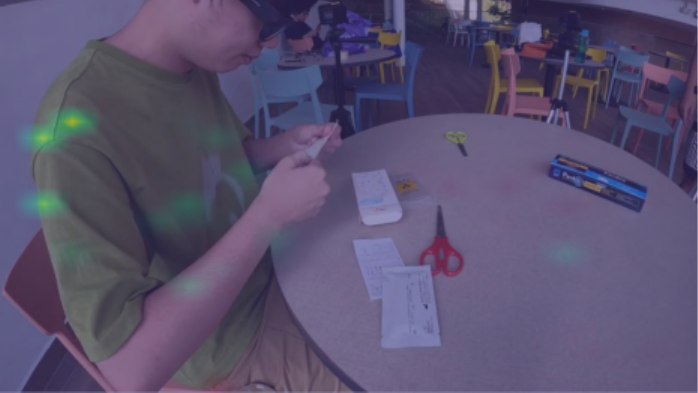}}} & 
{{\includegraphics[width=\cellwidthteaser cm, height=\cellwidthteaser cm]{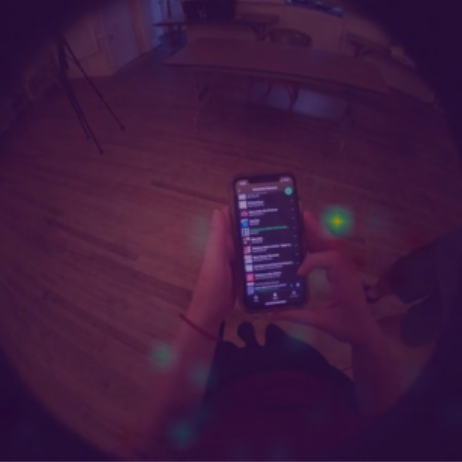}}} & 
{{\includegraphics[width=\cellwidthteaser cm, height=\cellwidthteaser cm]{assets/visualization/llava_visualization/gender_exo_img001014.pdf}}} & 
{{\includegraphics[width=\cellwidthteaser cm, height=\cellwidthteaser cm]{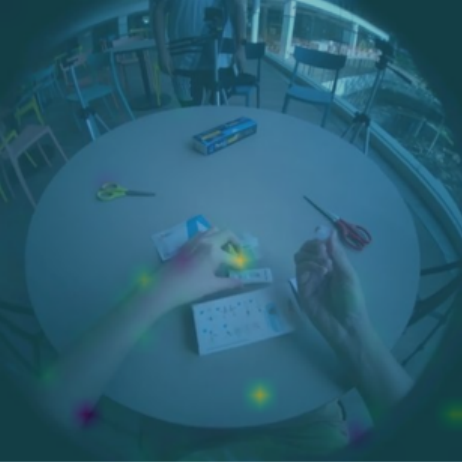}}} & 
{{\includegraphics[width=\cellwidthteaser cm, height=\cellwidthteaser cm]{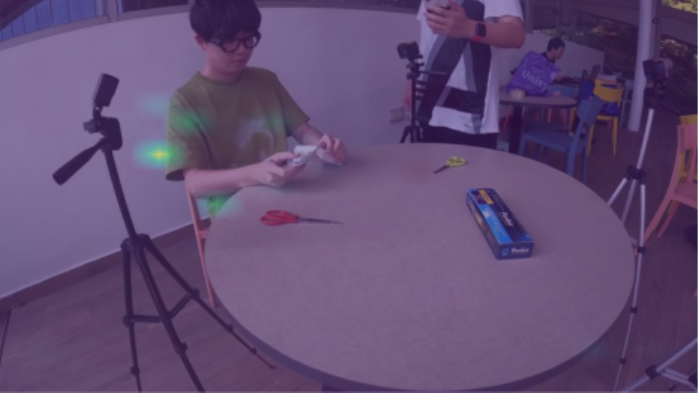}}}
\tabularnewline
% \midrule
\bottomrule
\end{tabular}
\end{center}
\vspace{-0.15in}
\caption{Attention Visualization of LLaVa model.}
\label{tab:llava_visualization}
\vspace{-0.15in}
\end{table}

\begin{table}[t]
\scriptsize
\begin{center}
\setlength{\tabcolsep}{1pt} % Default value: 6pt
\renewcommand{\arraystretch}{1.2}
% \footnotesize 
\begin{tabular}{cccccccc}
\toprule
 \multicolumn{2}{c}{mask ratio} & 0\% & 10\%  & 30\% & 50\%  & 70\% & 90\% 
\tabularnewline
\midrule
\multirow{2}{*}{Gender} & Exo & {{\includegraphics[width=\cellwidthteaser cm, height=\cellwidthteaser cm]{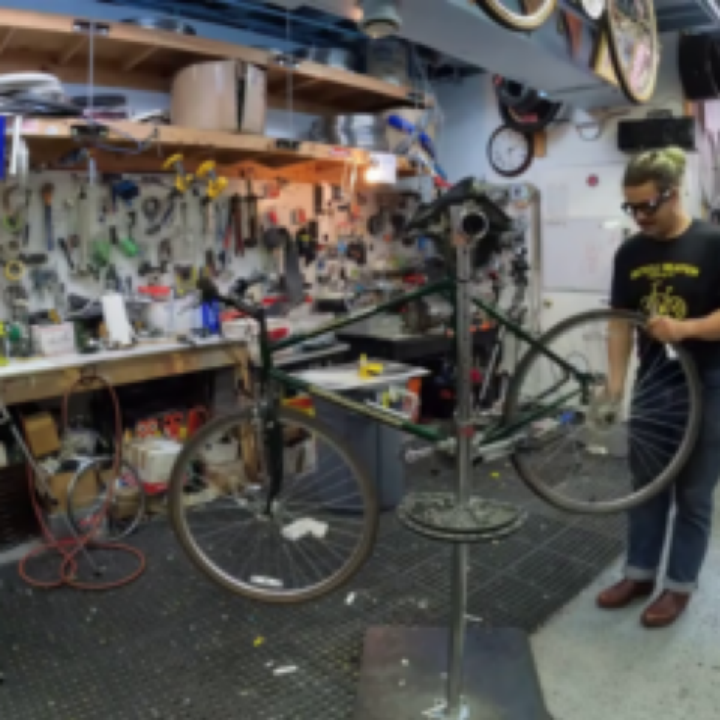}}} &
{{\includegraphics[width=\cellwidthteaser cm, height=\cellwidthteaser cm]{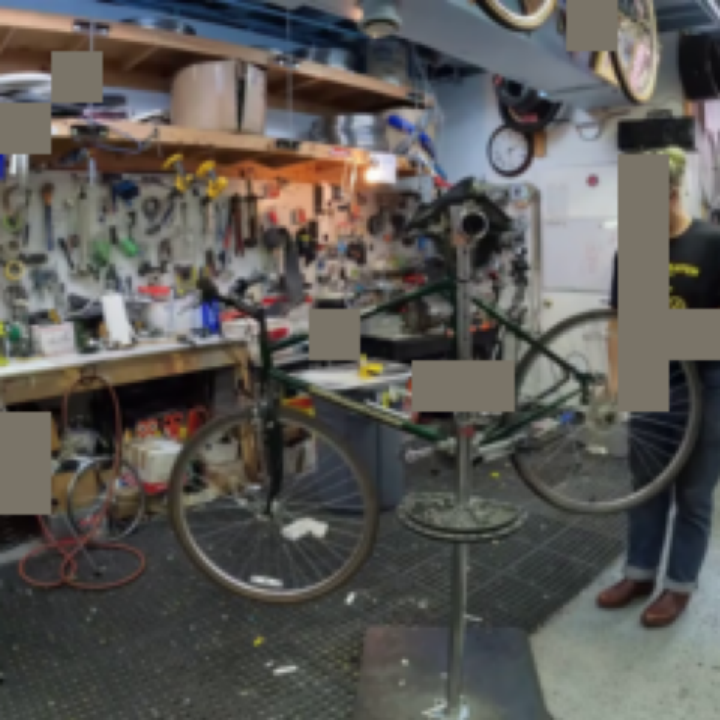}}} & 
{{\includegraphics[width=\cellwidthteaser cm, height=\cellwidthteaser cm]{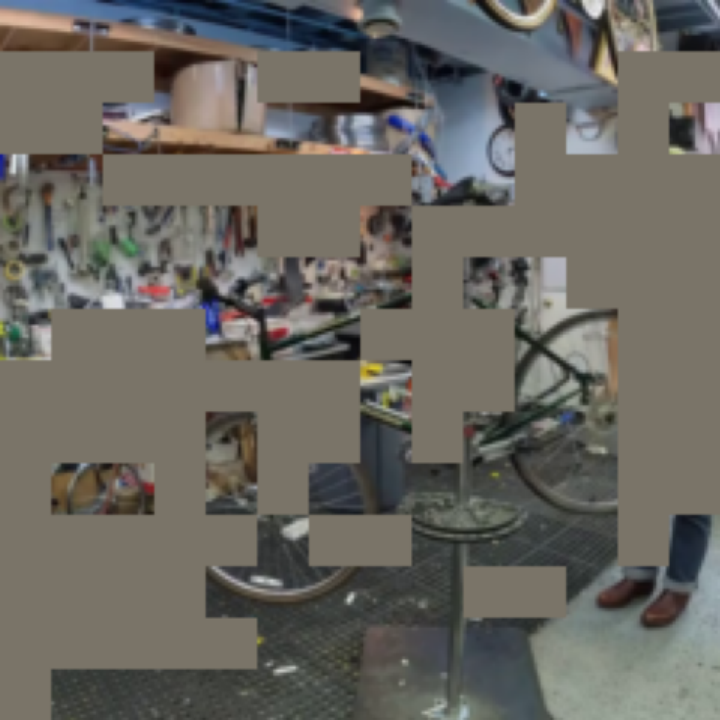}}} & 
{{\includegraphics[width=\cellwidthteaser cm, height=\cellwidthteaser cm]{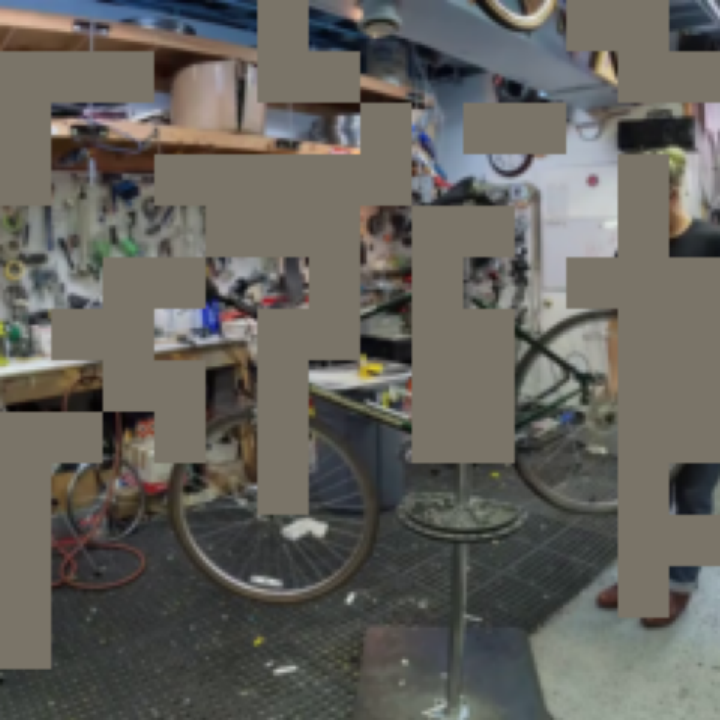}}} & 
{{\includegraphics[width=\cellwidthteaser cm, height=\cellwidthteaser cm]{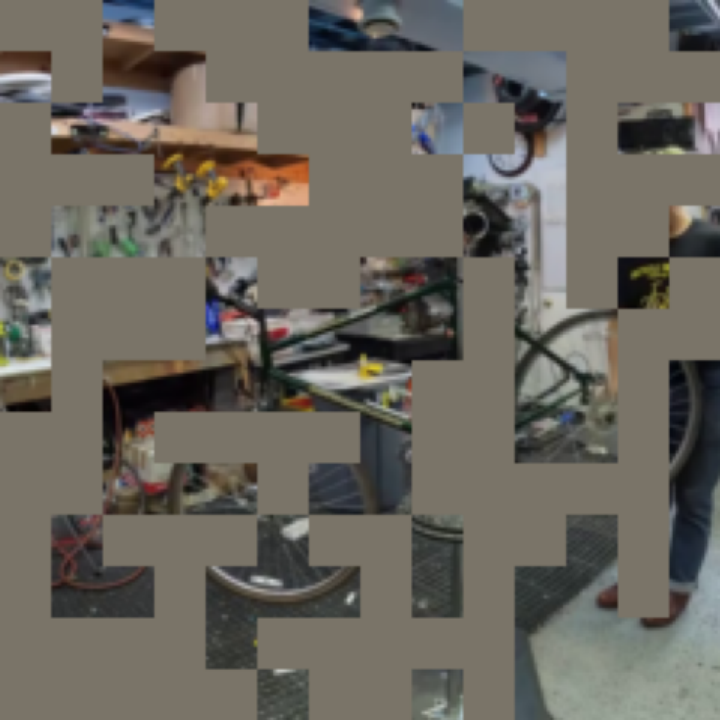}}} & 
{{\includegraphics[width=\cellwidthteaser cm, height=\cellwidthteaser cm]{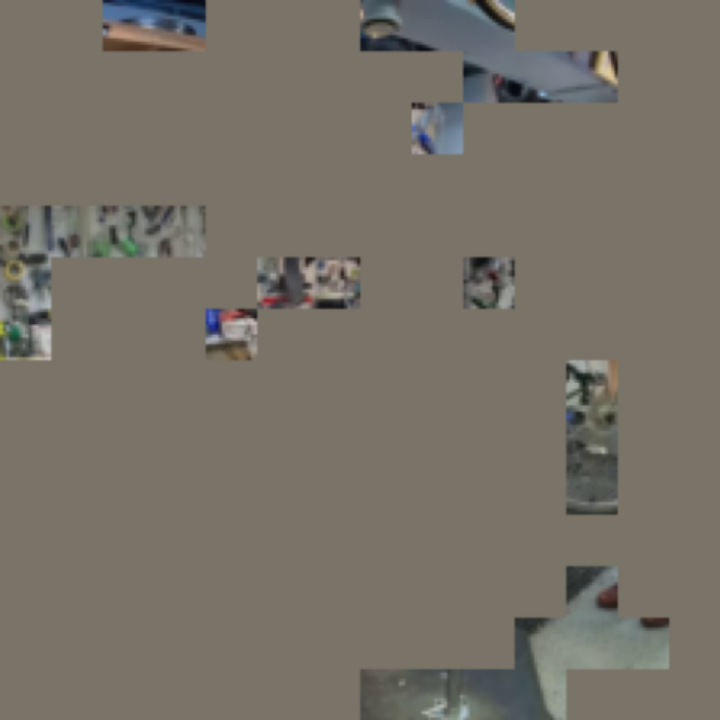}}} 
\tabularnewline
\cmidrule(r){2-8} 
 & Ego & {{\includegraphics[width=\cellwidthteaser cm, height=\cellwidthteaser cm]{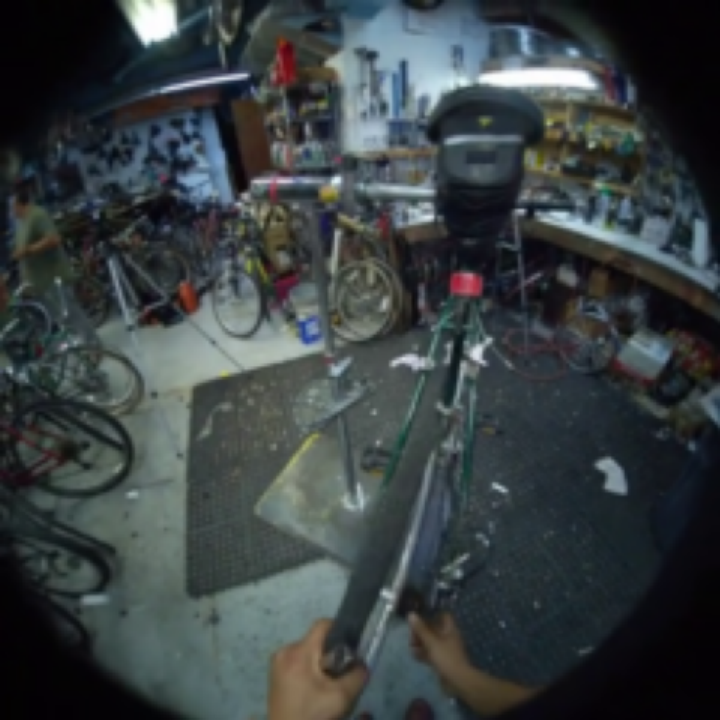}}} &
{{\includegraphics[width=\cellwidthteaser cm, height=\cellwidthteaser cm]{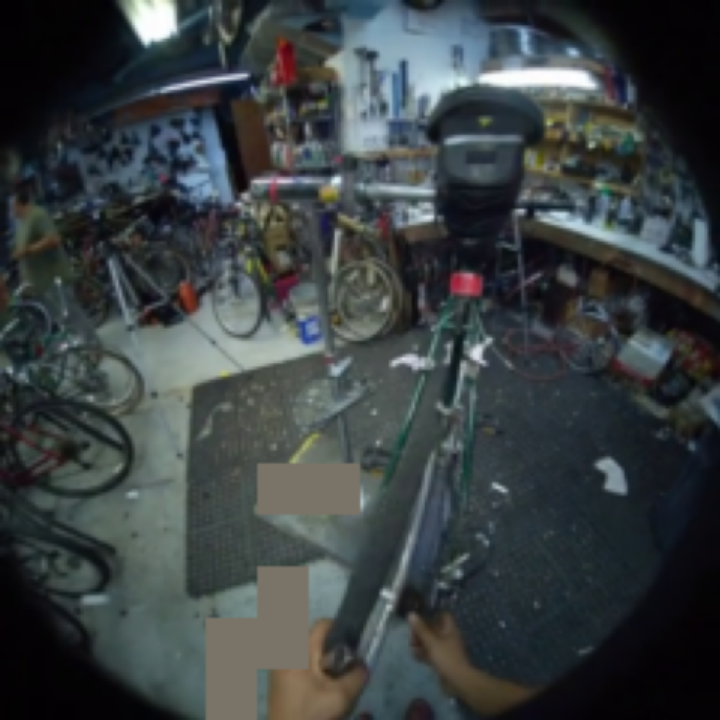}}} & 
{{\includegraphics[width=\cellwidthteaser cm, height=\cellwidthteaser cm]{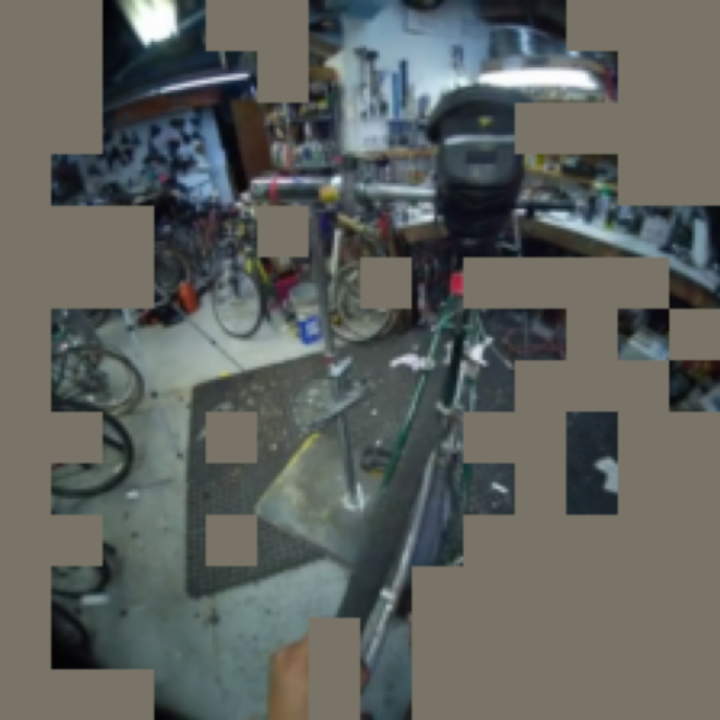}}} & 
{{\includegraphics[width=\cellwidthteaser cm, height=\cellwidthteaser cm]{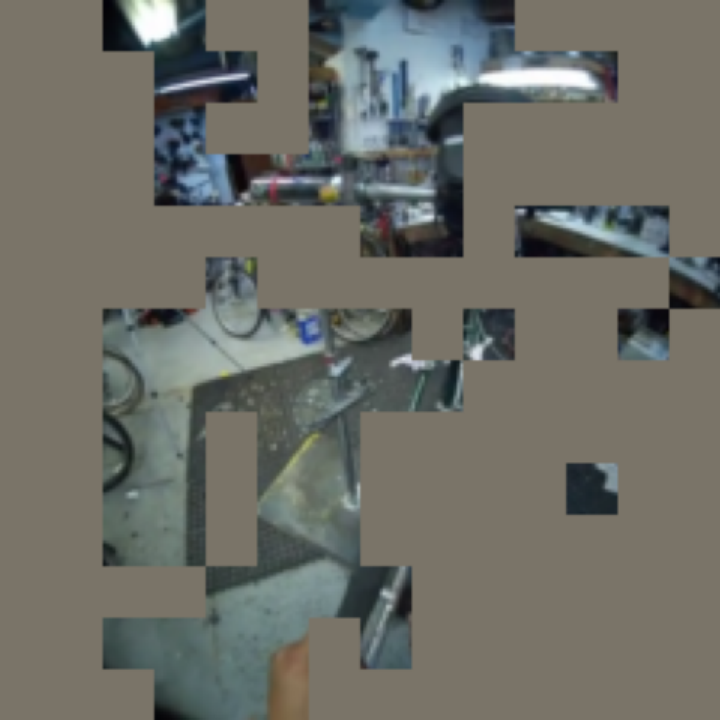}}} & 
{{\includegraphics[width=\cellwidthteaser cm, height=\cellwidthteaser cm]{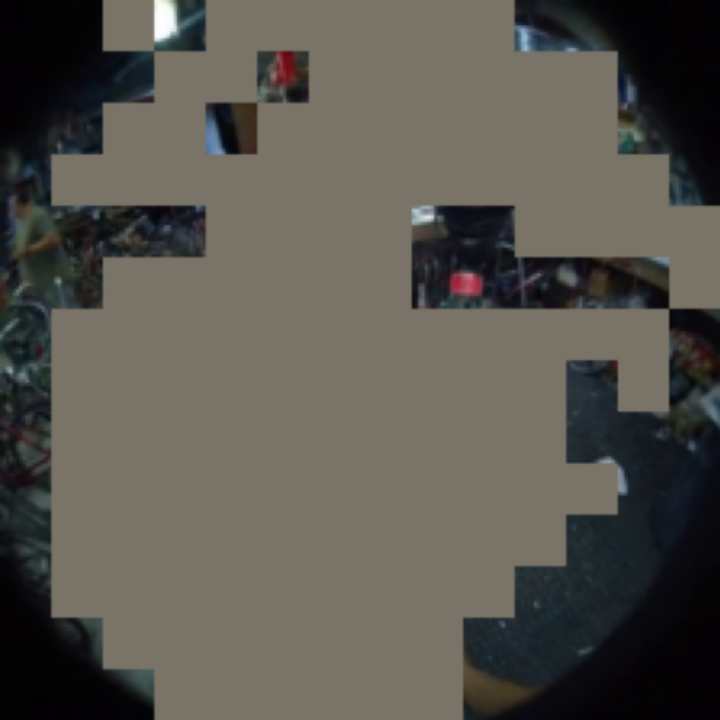}}} & 
{{\includegraphics[width=\cellwidthteaser cm, height=\cellwidthteaser cm]{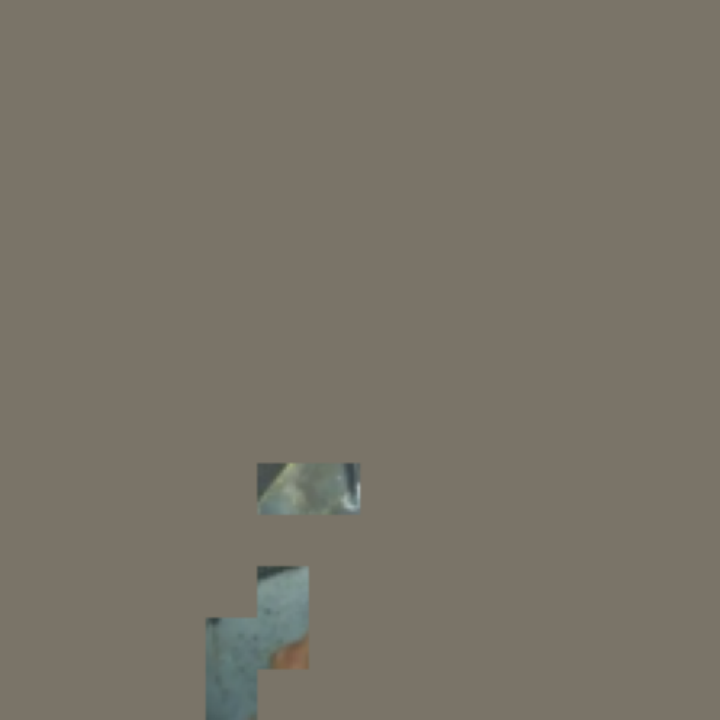}}} 
\tabularnewline
\midrule
\multirow{2}{*}{Race} & Exo & {{\includegraphics[width=\cellwidthteaser cm, height=\cellwidthteaser cm]{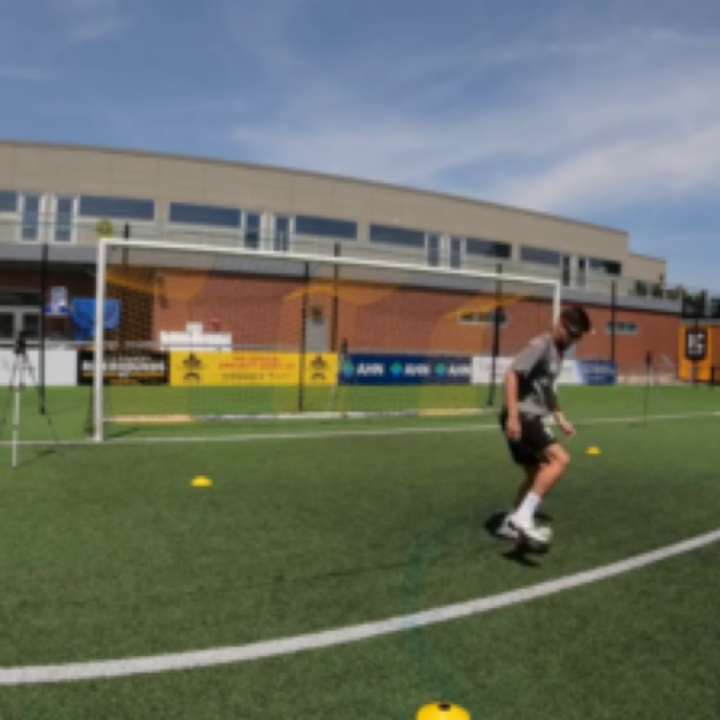}}} &
{{\includegraphics[width=\cellwidthteaser cm, height=\cellwidthteaser cm]{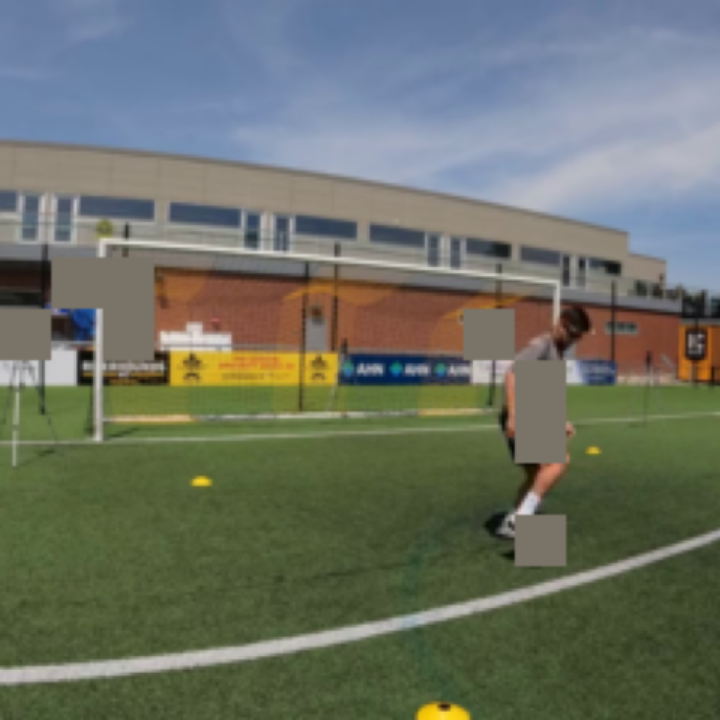}}} & 
{{\includegraphics[width=\cellwidthteaser cm, height=\cellwidthteaser cm]{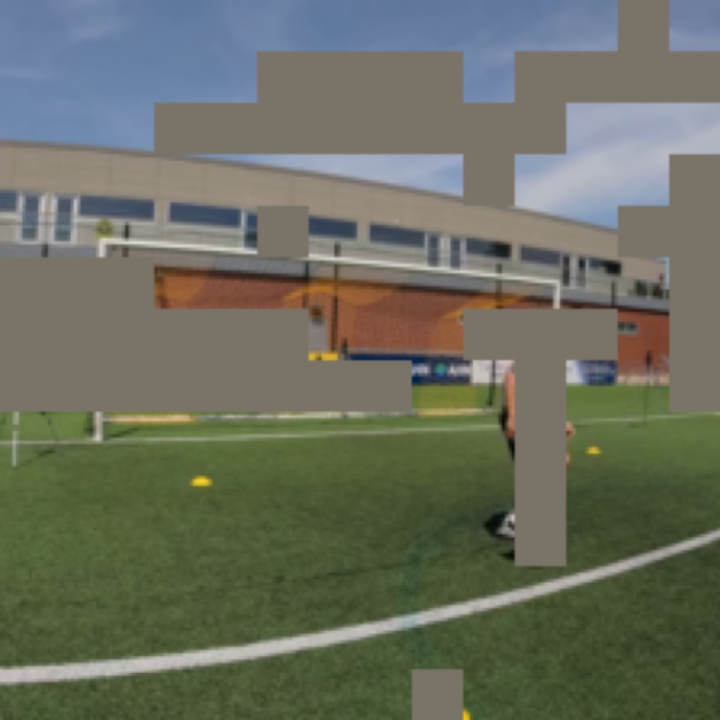}}} & 
{{\includegraphics[width=\cellwidthteaser cm, height=\cellwidthteaser cm]{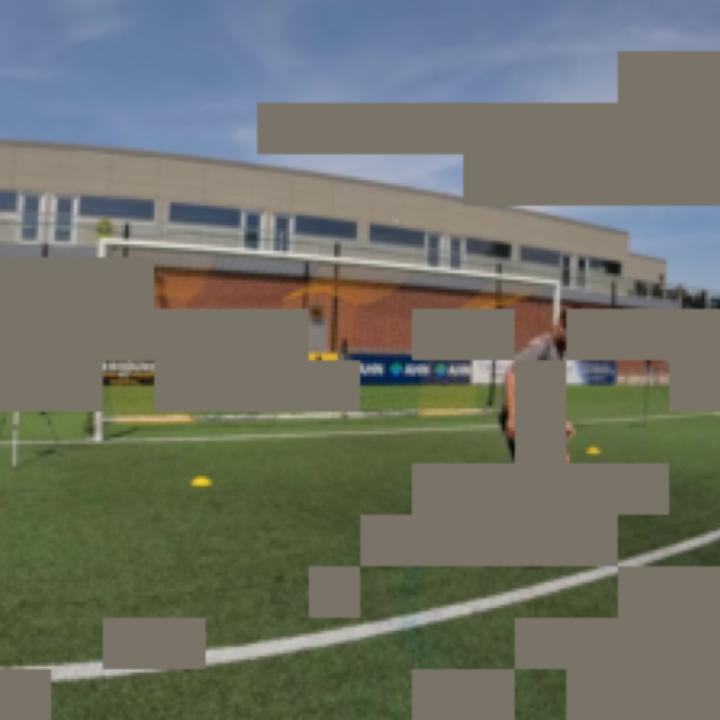}}} & 
{{\includegraphics[width=\cellwidthteaser cm, height=\cellwidthteaser cm]{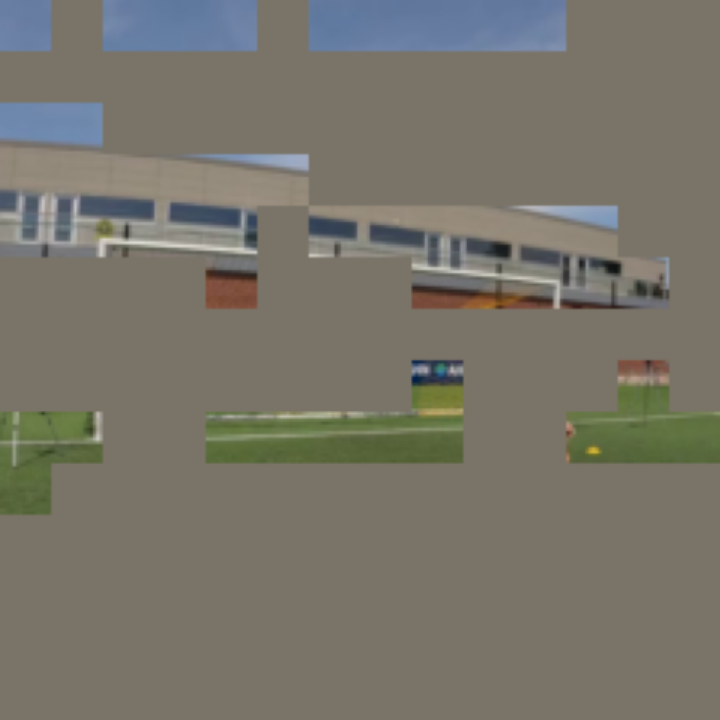}}} & 
{{\includegraphics[width=\cellwidthteaser cm, height=\cellwidthteaser cm]{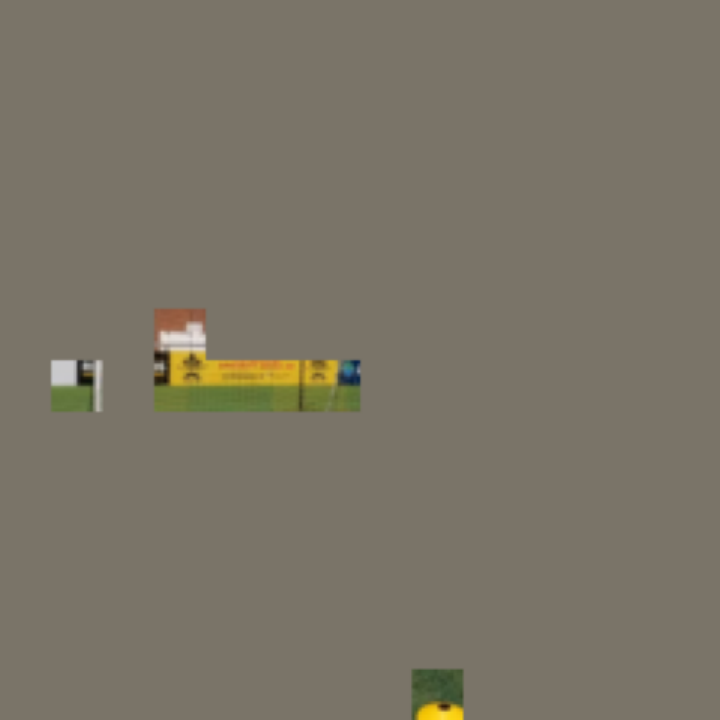}}} 
\tabularnewline
\cmidrule(r){2-8} 
 & Ego & {{\includegraphics[width=\cellwidthteaser cm, height=\cellwidthteaser cm]{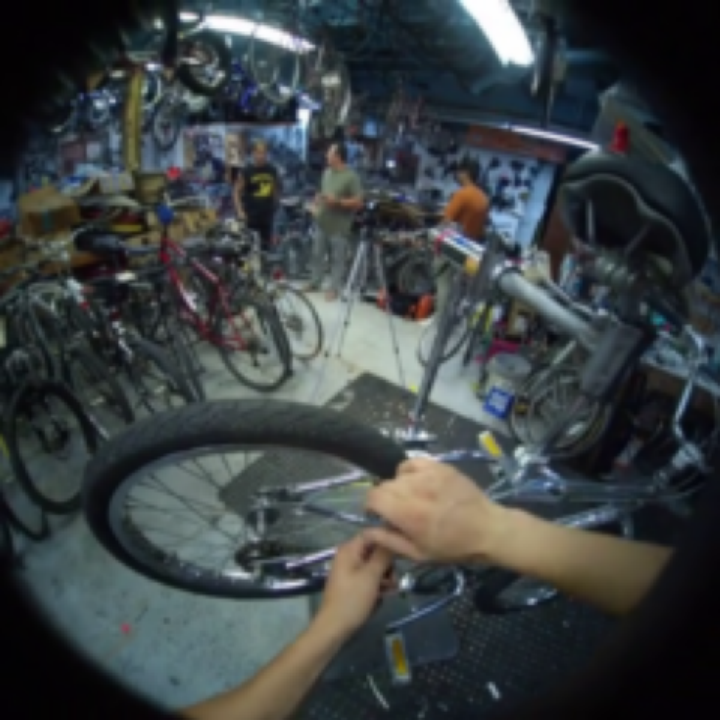}}} &
{{\includegraphics[width=\cellwidthteaser cm, height=\cellwidthteaser cm]{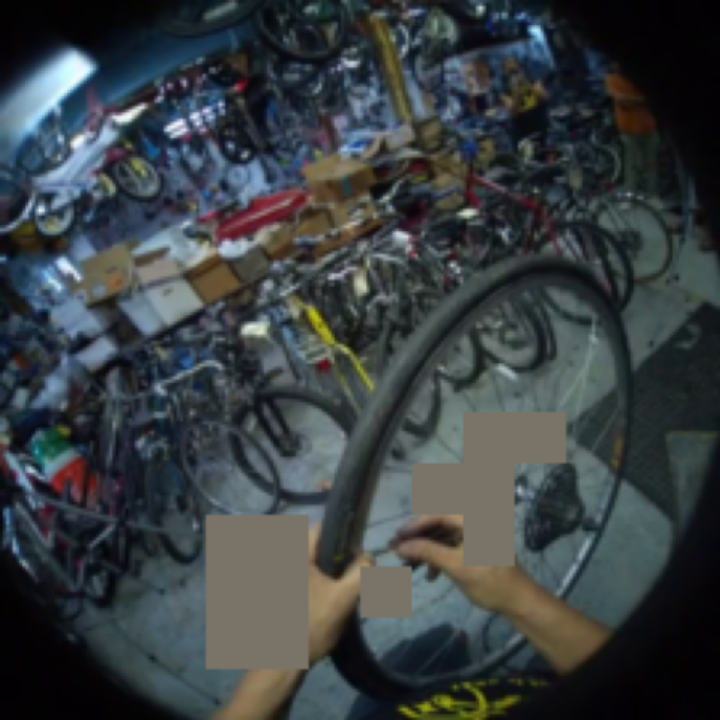}}} & 
{{\includegraphics[width=\cellwidthteaser cm, height=\cellwidthteaser cm]{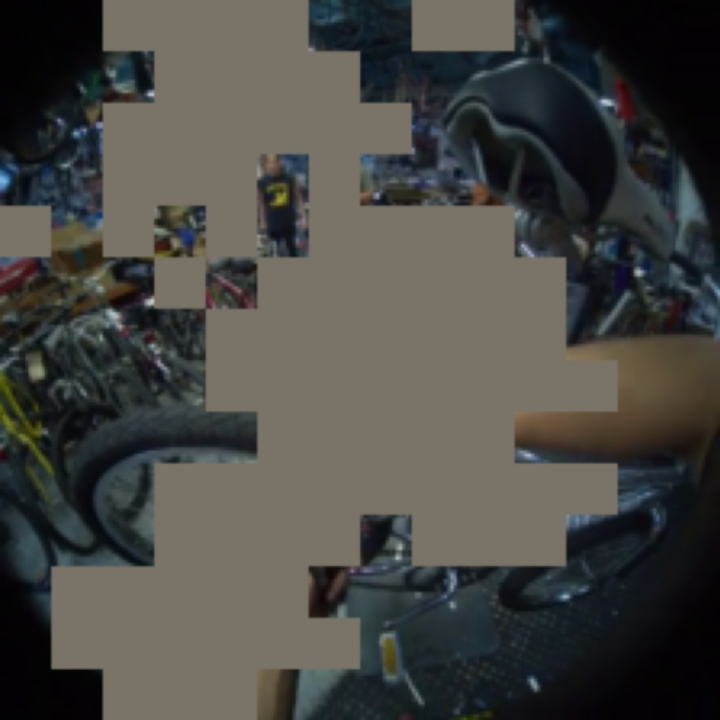}}} & 
{{\includegraphics[width=\cellwidthteaser cm, height=\cellwidthteaser cm]{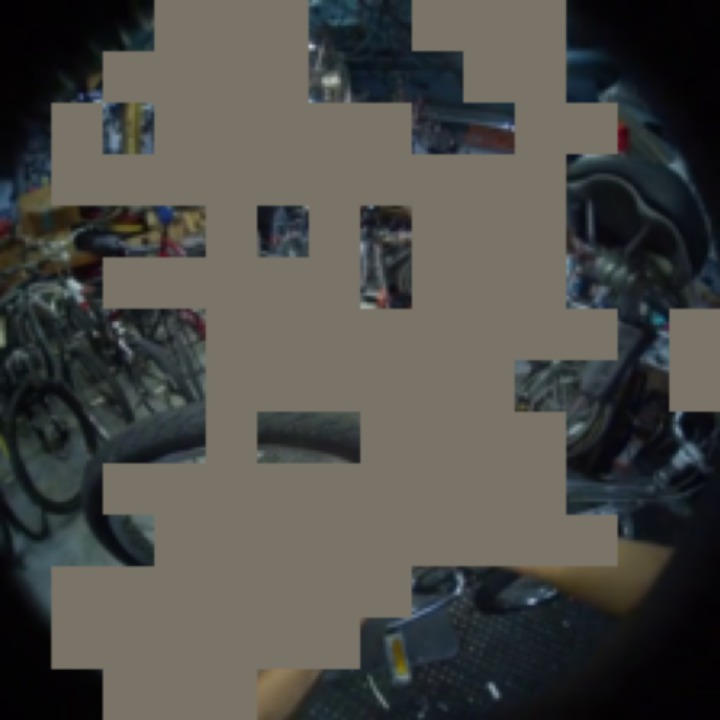}}} & 
{{\includegraphics[width=\cellwidthteaser cm, height=\cellwidthteaser cm]{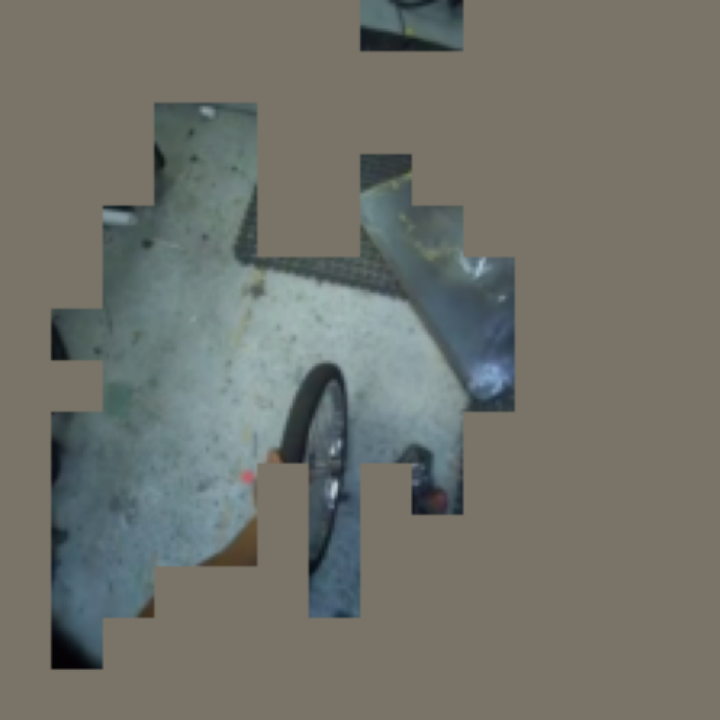}}} & 
{{\includegraphics[width=\cellwidthteaser cm, height=\cellwidthteaser cm]{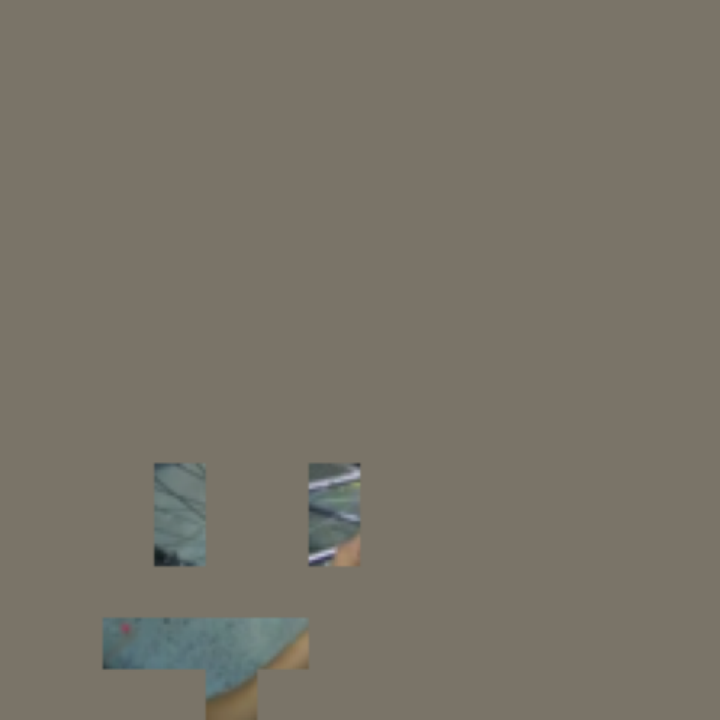}}} 
\tabularnewline
% \midrule
\bottomrule
\end{tabular}
\end{center}
\vspace{-0.1in}
\caption{Progressive masking of ego- and exo-video frames.}
\label{tab:progessive_masking}
\vspace{-0.1in}
\end{table}

\textbf{What leaks the privacy in the egocentric videos?} We visualize the attention of LLaVA when it makes the prediction in Table~\ref{tab:llava_visualization}. To further understand which patches contribute most to the prediction of privacy properties, we introduce a progressive masking method that incrementally masks the most important patches, as shown in Table~\ref{tab:progessive_masking}. We refer to  \cref{app:p_masking} for details of this method. Both visualizations reveal that significant attention is given to the wearer’s hand or other biometric markers.

\section{Conclusion}
\label{sec:conclusion}

In this work, we introduced \ours{}, a multidimensional benchmark of privacy in egocentric computer vision. By exploring demographic, individual, and situational privacy issues, we demonstrated that privacy information about the camera wearer can be extracted from first-person video data, even with off-the-shelf models in zero-shot. We proposed a retrieval-augmented attack, which further amplifies these threats by linking egocentric and exocentric footage of the same subjects. These results highlight the urgent need for privacy-preserving techniques in wearable cameras. We hope \ours{} will drive future research on safeguarding privacy in egocentric vision while maintaining its utility.

\clearpage
\section*{Acknowledgments}
This work was partially funded by NSF awards IIS-2303153 and NAIRR-240300, the NVIDIA Academic grant, and a gift from Qualcomm. We also acknowledge the NRP Nautilus cluster, used for some of the experiments discussed above.

\section*{Impact Statement}
This research reveals a significant vulnerability in wearable camera systems, demonstrating that egocentric privacy attacks can be effectively executed even using readily available, unmodified models. Although the introduced privacy attack methods, such as \emph{RAA}, are designed as red-teaming instruments aimed at enhancing privacy defenses, there exists a concerning potential for their misuse in unauthorized mass surveillance. Consequently, our findings highlight an urgent need for the development and implementation of robust privacy safeguards and proactive intervention mechanisms to mitigate risks associated with wearable technology. Furthermore, as \ours{} builds upon Ego-Exo4D and Charades-Ego, it inherits their imbalances in geographic, gender, ethnic, and age representation, which raise concerns about the fairness problem. This emphasizes the need for future efforts to curate more equitable datasets in egocentric vision and privacy research, which will be the next step of our work.

% % In the unusual situation where you want a paper to appear in the
% % references without citing it in the main text, use \nocite
% \nocite{langley00}

\bibliography{main}
\bibliographystyle{icml2025}

%%%%%%%%%%%%%%%%%%%%%%%%%%%%%%%%%%%%%%%%%%%%%%%%%%%%%%%%%%%%%%%%%%%%%%%%%%%%%%%
%%%%%%%%%%%%%%%%%%%%%%%%%%%%%%%%%%%%%%%%%%%%%%%%%%%%%%%%%%%%%%%%%%%%%%%%%%%%%%%
% APPENDIX
%%%%%%%%%%%%%%%%%%%%%%%%%%%%%%%%%%%%%%%%%%%%%%%%%%%%%%%%%%%%%%%%%%%%%%%%%%%%%%%
%%%%%%%%%%%%%%%%%%%%%%%%%%%%%%%%%%%%%%%%%%%%%%%%%%%%%%%%%%%%%%%%%%%%%%%%%%%%%%%
\newpage
\appendix
\onecolumn

\counterwithin{figure}{section}
\counterwithin{table}{section}
\numberwithin{equation}{section}

% \section{You \emph{can} have an appendix here.}

% You can have as much text here as you want. The main body must be at most $8$ pages long.
% For the final version, one more page can be added.
% If you want, you can use an appendix like this one.  

% The $\mathtt{\backslash onecolumn}$ command above can be kept in place if you prefer a one-column appendix, or can be removed if you prefer a two-column appendix.  Apart from this possible change, the style (font size, spacing, margins, page numbering, etc.) should be kept the same as the main body.

\section{Dataset}
\label{app:dataset}

\begin{figure}
    \centering
    \includegraphics[width=0.75\linewidth]{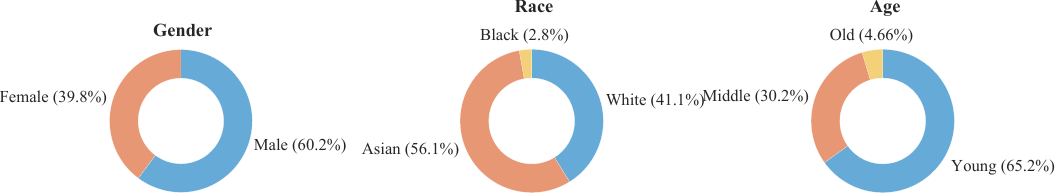}
    \caption{Distributions of demographic labels in \ours~(ID).}
    \label{fig:stats}
\end{figure}
\begin{figure}
    \centering
    \includegraphics[width=0.75\linewidth]{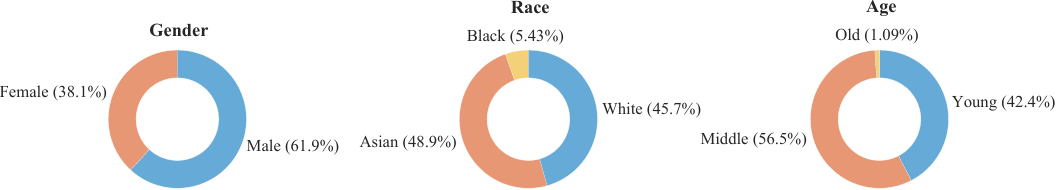}
    \caption{Distributions of demographic labels in \ours~(OOD).}
    \label{fig:stats}
\end{figure}

\paragraph{Data sources.}
We build our \ours{} upon two prior datasets with egocentric and exocentric annotation---Ego-Exo4D~\cite{grauman2024ego} and Charades-Ego ~\cite{sigurdsson2018charades}.
Ego-Exo4D comprises paired egocentric and exocentric videos capturing skilled activities performed by 740 participants across more than 100 distinct scenes in 13 cities worldwide. The dataset's diversity and extensive annotations enable privacy research at an unprecedented scale, making this study feasible for the first time. In Ego-Exo4D, each recording contains one or multiple trials (``takes'') of an activity, with each take spanning 2.6 minutes on average. 
The dataset was released with labels of participant IDs associated with each video as well as self-reported demographics of some of the participants, making it an ideal candidate for studying privacy in egocentric vision. Ego-Exo4D dataset also provides redundant exocentric recordings, where each egocentric video is paired 4 exocentric view footage. Following the official dataset split, each participant is assigned exclusively to one of the train/val/test sets, preventing leakage of identity or demographic information in learning the attack models. The other dataset we adopt for \ours{} is the Charades-Ego dataset. Charades-Ego is a dataset featuring 7,860 videos of daily indoor activities recorded from both third-person and first-person perspectives, comprising 68,536 temporal annotations across 157 action classes. Both videos possess paired egocentric and exocentric videos fulfilling the first requirement. To further satisfy the second requirement, we undergo an annotation process to label each identity of its gender, race and age. We note here that both the Ego-Exo4D and Charades-Ego dataset comes with identity labels. This is beneficial as it can reduce not only the annotation for identity but also the annotation cost of demographics for each video (since we can now annotate at the identity level).

\paragraph{Annotation Process.}
All videos and participant data used in this study come from publicly released datasets where participants consented to data collection. For participants who did not voluntarily disclose demographic information, we use crowd-sourced annotations of \emph{perceived} attributes based on their video appearances. 
We employ Amazon Mechanical Turk for demographic annotation. For each identity, we display 3 to 4 (depending on the availability) \emph{exocentric} videos to the annotator and request the annotator to answer three multi-choice questions regarding gender, race and age respectively. For each identity, we hire five Turker to annotate and filter any annotation with confidence less than 80\%.
These perceived demographics do not necessarily reflect individuals’ self-identities. All collected data are used solely for academic research on privacy risks in egocentric vision, and we take measures to safeguard the confidentiality of participant information.

\begin{figure}
    \centering
    \includegraphics[width=0.9\linewidth]{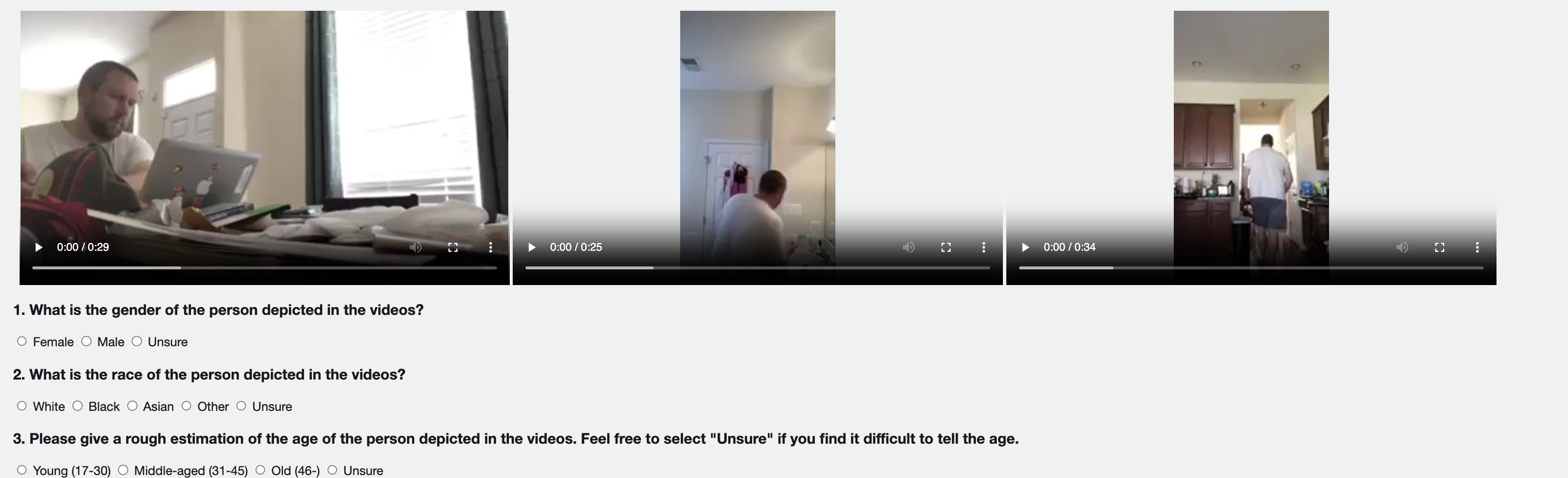}
    \caption{Amazon Mechanical Turk web user interface for demographic annotation.}
    \label{fig:enter-label}
\end{figure}

\section{Identity-level Privacy Attacks (Capability \textcircled{4})}
\label{ability4}
 We repeat the \emph{demographic} privacy attacks of \cref{tab:results_demographic}, but assume the additional capability \tikz[baseline=(char.base)]{
            \node[shape=circle,draw,inner sep=0.5pt] (char) {\textbf{4}};} of attackers, i.e. the ability to ascertain whether two egocentric videos share the same identity. We expect the attacker to further improve the attack performance with this extra information, which is the case for gender egocentric and all exocentric videos, as shown in Table \ref{tab:cap4}. However, the performance on egocentric age and race surprisingly drops. 

\begin{table*}[t]
    \centering
\resizebox{0.95\linewidth}{!}{%
\scriptsize
\begin{tabular}{@{}lcccc|cccr|cccr|cccr@{}}
\toprule

% & \textbf{OOD} & \multicolumn{2}{c|}{\textbf{Capability}} & \multicolumn{4}{c|}{\textbf{Gender}} & \multicolumn{4}{c|}{\textbf{Race}} & \multicolumn{4}{c}{\textbf{Age}} \\
% % \multicolumn{2}{c|}{\multirow{-2}{*}{\textit{\begin{tabular}[c]{@{}c@{}}Ego-Exo4D\\ (ID)\end{tabular}}}} 
% & (Charades-Ego) & \tikz[baseline=(char.base)]{\node[shape=circle,draw,inner sep=0.5pt] (char) {\textbf{1}}} & \tikz[baseline=(char.base)]{\node[shape=circle,draw,inner sep=0.5pt] (char) {\textbf{2}}}
% & 
% {\color[HTML]{9B9B9B} Exo} & Ego & RAA (+ \tikz[baseline=(char.base)]{\node[shape=circle,draw,inner sep=0.5pt] (char) {\textbf{3}}}) & \multicolumn{1}{c|}{$\Delta$} 
% & {\color[HTML]{9B9B9B} Exo} & Ego & RAA (+ \tikz[baseline=(char.base)]{\node[shape=circle,draw,inner sep=0.5pt] (char) {\textbf{3}}}) & \multicolumn{1}{c|}{$\Delta$} 
% & {\color[HTML]{9B9B9B} Exo} & Ego & RAA (+ \tikz[baseline=(char.base)]{\node[shape=circle,draw,inner sep=0.5pt] (char) {\textbf{3}}}) & \multicolumn{1}{c}{$\Delta$} \\ 
% \midrule
% \rowcolor[HTML]{EFEFEF} 
% Random Chance & & & & \multicolumn{3}{c}{\cellcolor[HTML]{EFEFEF}50.00} & \multicolumn{1}{c|}{\cellcolor[HTML]{EFEFEF}-} & \multicolumn{3}{c}{\cellcolor[HTML]{EFEFEF}33.33} & \multicolumn{1}{c|}{\cellcolor[HTML]{EFEFEF}-} & \multicolumn{3}{c}{\cellcolor[HTML]{EFEFEF}33.33} & \multicolumn{1}{c}{\cellcolor[HTML]{EFEFEF}-} \\ 

& \textbf{OOD} & \multicolumn{3}{c|}{\textbf{Capability}} & \multicolumn{4}{c|}{\textbf{Gender}} & \multicolumn{4}{c|}{\textbf{race}} & \multicolumn{4}{c}{\textbf{Age}} \\
& (Charades-Ego) & \tikz[baseline=(char.base)]{\node[shape=circle,draw,inner sep=0.5pt] (char) {\textbf{1}}} & \tikz[baseline=(char.base)]{\node[shape=circle,draw,inner sep=0.5pt] (char) {\textbf{2}}} & \tikz[baseline=(char.base)]{\node[shape=circle,draw,inner sep=0.5pt] (char) {\textbf{4}}}
& 
{\color[HTML]{9B9B9B} Exo} & Ego & RAA (+ \tikz[baseline=(char.base)]{\node[shape=circle,draw,inner sep=0.5pt] (char) {\textbf{3}}}) & \multicolumn{1}{c|}{$\Delta$} 
& {\color[HTML]{9B9B9B} Exo} & Ego & RAA (+ \tikz[baseline=(char.base)]{\node[shape=circle,draw,inner sep=0.5pt] (char) {\textbf{3}}}) & \multicolumn{1}{c|}{$\Delta$} 
& {\color[HTML]{9B9B9B} Exo} & Ego & RAA (+ \tikz[baseline=(char.base)]{\node[shape=circle,draw,inner sep=0.5pt] (char) {\textbf{3}}}) & \multicolumn{1}{c}{$\Delta$} \\ 
\midrule
\rowcolor[HTML]{EFEFEF} 
Random Chance & & & &  & & {\cellcolor[HTML]{EFEFEF}50.00} & & \multicolumn{1}{c|}{\cellcolor[HTML]{EFEFEF}-} &  & {\cellcolor[HTML]{EFEFEF}33.33} & & \multicolumn{1}{c|}{\cellcolor[HTML]{EFEFEF}-} & & {\cellcolor[HTML]{EFEFEF}33.33} & & \multicolumn{1}{c}{\cellcolor[HTML]{EFEFEF}-} \\ 
\midrule
\multirow{4}{*}{$\text{CLIP}_{\text{H/14}}$}  & \xmark & \cmark & \xmark & \cmark & {\color[HTML]{9B9B9B} 84.97} & 62.07 & 71.26 & {\color[HTML]{009901} 9.19} & {\color[HTML]{9B9B9B} 62.84} & 59.17 & 62.13 & {\color[HTML]{009901} 2.96} & {\color[HTML]{9B9B9B} 73.03} & 67.63 & 73.99 & {\color[HTML]{009901} 6.36}\\
&\xmark & \xmark & \cmark & \cmark & {\color[HTML]{9B9B9B} 89.54} & 69.54 & 77.59 & {\color[HTML]{009901} 8.05} & {\color[HTML]{9B9B9B} 75.68} & 70.41 & 72.19 & {\color[HTML]{009901} 1.78} & {\color[HTML]{9B9B9B} 74.34} & 76.30 & 82.08 & {\color[HTML]{009901} 5.78}\\
&\cmark & \cmark & \xmark & \cmark & {\color[HTML]{9B9B9B} 93.02} & 76.19 & 79.43 & \multicolumn{1}{r|}{{\color[HTML]{009901} 3.24}} & {\color[HTML]{9B9B9B} 68.60} & 58.33 & 63.71 & \multicolumn{1}{r|}{{\color[HTML]{009901} 5.38}} & {\color[HTML]{9B9B9B} 54.65} & 20.24 & 27.00 & {\color[HTML]{009901} 6.76} \\
& \cmark& \xmark & \cmark & \cmark & {\color[HTML]{9B9B9B} 77.38} & 55.68 & 70.01 & \multicolumn{1}{r|}{{\color[HTML]{009901} 14.39}} & {\color[HTML]{9B9B9B} 86.08} & 66.79 & 77.03 & \multicolumn{1}{r|}{{\color[HTML]{009901} 10.24}} & {\color[HTML]{9B9B9B} 28.56} & 28.20 & 29.35 & {\color[HTML]{009901} 1.15} \\
\midrule
\multirow{2}{*}{EgoVLP v2} & \xmark& \xmark & \cmark & \cmark & {\color[HTML]{9B9B9B} 89.54} & 71.84 & 77.57 & {\color[HTML]{009901} 5.73} & {\color[HTML]{9B9B9B} 77.70} & 72.19 & 78.70 & {\color[HTML]{009901} 6.51} & {\color[HTML]{9B9B9B}  75.00} & 78.03 & 78.03 & 0.00 \\
& \cmark& \xmark & \cmark  & \cmark & {\color[HTML]{9B9B9B} 78.16} & 55.32 & 68.02 & \multicolumn{1}{r|}{{\color[HTML]{009901} 12.70}} & {\color[HTML]{9B9B9B} 77.32} & 61.77 & 73.54 & \multicolumn{1}{r|}{{\color[HTML]{009901} 11.77}} & {\color[HTML]{9B9B9B} 29.20} & 28.20 & 28.57 & {\color[HTML]{009901} 0.37}\\
% VideoMAE$_{\text{B/14}}$ & \xmark & \cmark & \cmark \\
% VideoMAE$_{\text{L/14}}$ & \xmark & \cmark  & \cmark \\
\midrule
\multirow{2}{*}{$\text{LLaVA-1.5}_{\text{7B}}$} &\xmark & \cmark & \xmark & \cmark & {\color[HTML]{9B9B9B}96.08} & 71.26 & 72.99 & {\color[HTML]{009901} 1.73} & {\color[HTML]{9B9B9B} 67.57} & 52.66 & 66.27 & {\color[HTML]{009901} 13.61} & {\color[HTML]{9B9B9B} 79.61} & 76.30 & 77.46 & {\color[HTML]{009901} 1.16}\\
&\cmark & \cmark & \xmark & \cmark  & {\color[HTML]{9B9B9B} 92.71} & 71.43 & 77.59 & \multicolumn{1}{r|}{{\color[HTML]{009901} 6.16}} & {\color[HTML]{9B9B9B} 72.33} & 52.90 & 66.50 & \multicolumn{1}{r|}{{\color[HTML]{009901} 13.60}} & {\color[HTML]{9B9B9B} 52.88} & 37.48 & 41.48 & {\color[HTML]{009901} 4.00}\\
\multirow{2}{*}{$\text{LLaVA-1.5}_{\text{13B}}$} & \xmark& \cmark & \xmark & \cmark & {\color[HTML]{9B9B9B} 97.39} & 67.24 & 74.14 & {\color[HTML]{009901} 6.90} & {\color[HTML]{9B9B9B} 70.95} & 59.76 & 64.50 & {\color[HTML]{009901} 4.74} & {\color[HTML]{9B9B9B} 78.95} & 60.12 & 76.88 & {\color[HTML]{009901} 16.76} \\
& \cmark& \cmark & \xmark & \cmark & {\color[HTML]{9B9B9B} 95.35} & 71.43 & 78.56 & \multicolumn{1}{r|}{{\color[HTML]{009901} 7.13}} & {\color[HTML]{9B9B9B} 70.24} & 52.69 & 62.42 & \multicolumn{1}{r|}{{\color[HTML]{009901} 9.73}} & {\color[HTML]{9B9B9B} 52.88} & 36.72 & 42.38 & {\color[HTML]{009901} 5.66} \\
\midrule
\multirow{2}{*}{$\text{VideoLLaMA2}_{\text{7B}}$} & \xmark& \cmark & \xmark & \cmark & {\color[HTML]{9B9B9B} 98.04} & 77.01 & 80.46 & {\color[HTML]{009901} 3.45} & {\color[HTML]{9B9B9B} 77.03} & 60.36 & 74.56 & {\color[HTML]{009901} 14.20} & {\color[HTML]{9B9B9B} 56.58} & 42.77 & 52.60 & {\color[HTML]{009901} 9.83}  \\
&\cmark & \cmark & \xmark & \cmark & {\color[HTML]{9B9B9B} 92.85} & 72.56 & 78.39 & \multicolumn{1}{r|}{{\color[HTML]{009901} 5.83}} & {\color[HTML]{9B9B9B} 77.01} & 62.97 & 69.55 & \multicolumn{1}{r|}{{\color[HTML]{009901} 6.58}} & {\color[HTML]{9B9B9B} 67.92} & 57.11 & 59.49 & {\color[HTML]{009901} 2.38} \\
\multirow{2}{*}{$\text{VideoLLaMA2}_{\text{72B}}$} & \xmark& \cmark & \xmark & \cmark & {\color[HTML]{9B9B9B} 98.04} & 72.41 & 83.33 & {\color[HTML]{009901} 10.92} & {\color[HTML]{9B9B9B}  72.97} & 63.91 & 71.60 & {\color[HTML]{009901} 7.69} & {\color[HTML]{9B9B9B} 80.26} & 76.30 & 82.08 & {\color[HTML]{009901} 5.78}   \\
&\cmark & \cmark & \xmark & \cmark & {\color[HTML]{9B9B9B} 95.33} & 74.54 & 79.90 & \multicolumn{1}{r|}{{\color[HTML]{009901} 5.36}} & {\color[HTML]{9B9B9B} 77.92} & 68.22 & 70.35 & \multicolumn{1}{r|}{{\color[HTML]{009901} 2.13}} & {\color[HTML]{9B9B9B} 57.01} & 33.88 & 47.09 & {\color[HTML]{009901} 13.32}\\
\bottomrule
\end{tabular}%
}
    \caption{Results on \textbf{Demographic Privacy}. Accuracy is calculated on a \emph{per-identity} basis with the assumption of capability \tikz[baseline=(char.base)]{
            \node[shape=circle,draw,inner sep=0.5pt] (char) {\textbf{4}};}.}
    \label{tab:cap4}
\end{table*}

\section{Justification of Threat Model Capabilities}
\label{justification}

We discuss capabilities \tikz[baseline=(char.base)]{\node[shape=circle,draw,inner sep=0.5pt] (char) {\textbf{3}};} and \tikz[baseline=(char.base)]{\node[shape=circle,draw,inner sep=0.5pt] (char) {\textbf{4}};} and justify their necessity by illustrating their relevance to real-world scenarios. For capability \tikz[baseline=(char.base)]{\node[shape=circle,draw,inner sep=0.5pt] (char) {\textbf{3}};}, consider a case where the target individual is a student who shares egocentric videos online, and an adversary gains access to surveillance cameras in public areas of the student’s school. Capability \tikz[baseline=(char.base)]{\node[shape=circle,draw,inner sep=0.5pt] (char) {\textbf{4}};} is even more pervasive: here, the target posts multiple egocentric videos on social media, allowing an adversary to infer that all videos associated with the same account belong to a single individual. The objective of the adversary is then, given all the egocentric videos in the same account, infer the privacy attributes and information of the account owner. These examples highlight the practical relevance and necessity of these capabilities within our threat model.

% We here discuss the \tikz[baseline=(char.base)]{
%             \node[shape=circle,draw,inner sep=0.5pt] (char) {\textbf{3}};} and \tikz[baseline=(char.base)]{
%             \node[shape=circle,draw,inner sep=0.5pt] (char) {\textbf{4}};} capabilities and justify their necessity by demonstrating their relevance to real-world scenarios. For \tikz[baseline=(char.base)]{
%             \node[shape=circle,draw,inner sep=0.5pt] (char) {\textbf{3}};}, we can imagine a scenario where the target identity is a student posting egocentric videos on the internet and the adversary manages to access the surveillance cameras placed in the public areas of the identity's school. 
%             Capability \tikz[baseline=(char.base)]{
%             \node[shape=circle,draw,inner sep=0.5pt] (char) {\textbf{4}};}, would be even a more widely existent scenario, where the target identity posting numerous egocentric videos on the social media.Then the adversary deduces that all the videos under the same account belong to the same identity.

%             These examples underscore the relevance and necessity of these capabilities in the threat model under consideration.

\section{Details of Progressive Masking Method}
\label{app:p_masking}
In order to explore what features exactly in the video and frames that leaks the privacy information. We derive a progressive masking method that incrementally masks the most important patches.
Specifically, we initialize a mask with values between 0 and 1 and perform gradient ascent on the mask with respect to the privacy property prediction loss. By gradually increasing the number of masked patches and employing early stopping once a predefined threshold is reached, we constrain the masking process to reveal the patches most critical to the model’s decision.

\section{Biometric Classifier}

% It's suggested that prior methods have explored biometrics as cues for privacy inference and demonstrate promsing results. For exmaple, face-based classifier xxx FairFace, DEX, and hand-based biometric models  (Matkowski et al., 2019), (Matkowski et al., 2020)
% PolyU-IITD-v3: ~12k hand images with identity and race labels
% 11K Hands: ~11k hand images annotated with identity, age, gender, and race
% NTU-PI-v1: ~8k palm and dorsal hand images with identity, age, gender, and race labels
% CASIA: ~5.5k hand images labeled with identity

For the hand-based model, we trained a ResNet50 classifier on the publicly available 11K Hands dataset \cite{afifi201911k}, which contains gender and age labels (but lacks race annotation). During inference, hand regions were first detected and cropped from egocentric video frames using a YOLO-based hand detection model \cite{cansik2020yolo}. The resulting hand crops were then passed to the trained ResNet50 classifier to predict demographic attributes. To aggregate predictions across multiple hand regions, we applied majority voting.

For the face-based model, we employed the FairFace model, pretrained on the FairFace dataset \cite{karkkainenfairface}, together with RetinaFace for robust face detection \cite{deng2019retinafacesinglestagedenseface}. Faces were detected and cropped from exocentric video frames using RetinaFace, after which the cropped images were input to the FairFace model to predict demographic attributes such as gender and age. As shown in the second section of Table~\ref{tab:results_demographic}, these biometric methods perform substantially worse than even the zero-shot foundation model, likely due to a pronounced distribution gap between the small, curated datasets (hand/palm and face images) used for training and the more diverse, in-the-wild images in \ours.

\section{Effect of Temporal Modeling in Identity and Situational Privacy}
\label{app:addi_temporal_effect}
\begin{figure}[h]
\centering
\includegraphics[width=0.4\textwidth]{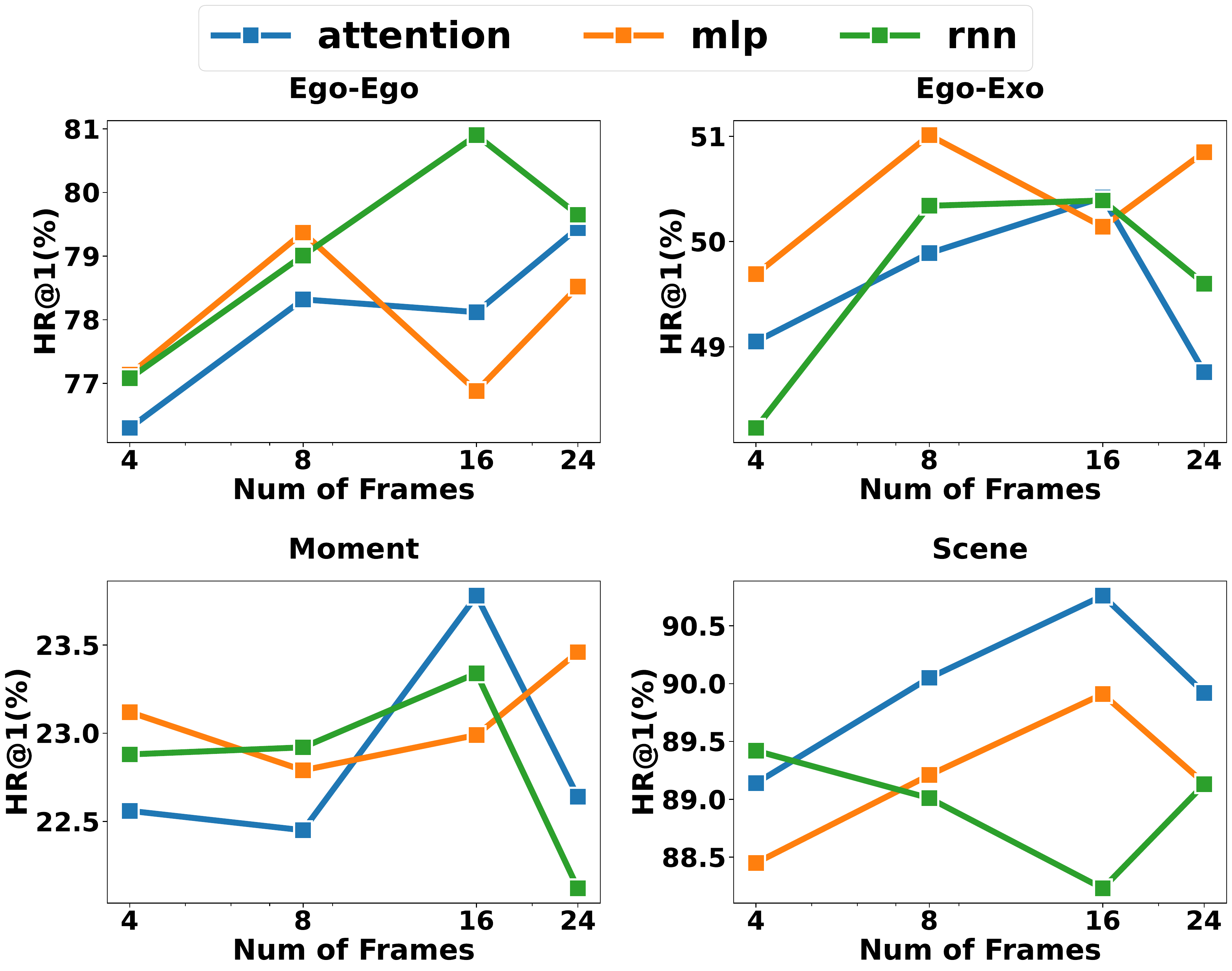}
\vspace{-0.2in}
\caption{Performance of Clip model with mlp, rnn and attention head on Identity and Situational Privacy.}
\label{figure_temporal_retrieval}
\end{figure}

We validate the observation in Section \ref{main_result} that temporal modeling is effective for adversary to reveal egocentric privacy, as shown in Figure~\ref{figure_temporal_retrieval}
\end{document}